\documentclass[10pt,twocolumn,letterpaper]{article}

\usepackage{cvpr}              % To produce the CAMERA-READY version
\usepackage{CJK}
\usepackage{multicol}
\usepackage{color}
\usepackage{xcolor}
\usepackage{colortbl}
\usepackage{bm}
\usepackage{extarrows}
\usepackage{float}
\usepackage{bbm}
\usepackage{booktabs}
\usepackage{multirow}
\usepackage{amsmath}
\usepackage{amssymb}
\usepackage{algorithm}
\usepackage{algorithmic}
\usepackage{makecell}
\usepackage{amsmath}
\usepackage{amssymb}
\usepackage{amsthm}
\usepackage[utf8]{inputenc}
\usepackage{amsfonts} % \mathbf (z) 符号
\usepackage{caption} 
\usepackage{newfloat}
\usepackage{listings}
\usepackage[accsupp]{axessibility} 

\usepackage{xcolor}
\definecolor{thmcolor}{HTML}{0201F5}

\newtheoremstyle{coloredthm}% name
  {3pt}%      Space above
  {3pt}%      Space below
  {\itshape}%         Body font
  {}%         Indent amount (empty = no indent, \parindent = para indent)
  {\bfseries\color{thmcolor}}% Thm head font (Bold AND Colored)
  {.}%        Punctuation after thm head
  {.5em}%     Space after thm head: " " = normal interword space;
  {}%         Thm head spec (can be left empty, meaning 'normal')

\theoremstyle{coloredthm}

\definecolor{cvprblue}{rgb}{0.21,0.49,0.74}
\definecolor{supplcolor}{HTML}{A32D26}

\definecolor{noise0color}{rgb}{0.949, 0.949, 1.0}
\definecolor{noise20color}{HTML}{FFF9E6} % 极淡橙
\definecolor{noise50color}{HTML}{EDF7ED} % 极淡绿
\definecolor{noise80color}{rgb}{0.949, 0.949, 1.0}
\usepackage[pagebackref,breaklinks,colorlinks,citecolor=brown]{hyperref}
\usepackage{thmtools}

\def\paperID{*****} % *** Enter the Paper ID here
\def\confName{CVPR}
\def\confYear{2026}

\def\modelname{\mbox{ConeSep} }
\newcommand{\statement}[1]{\noindent\textbf{#1}}

\newcommand\blfootnote[1]{%
  \begingroup
  \renewcommand\thefootnote{}\footnote{#1}%
  \addtocounter{footnote}{-1}%
  \endgroup
}
\title{ConeSep: Cone-based Robust Noise-Unlearning Compositional Network for Composed Image Retrieval}

\author{Zixu Li$^{1}$~~~~Yupeng Hu$^{1*}$\blfootnote{~corresponding authors}~~~~Zhiwei Chen$^1$~~~~Mingyu Zhang$^{1}$~~~~Zhiheng Fu$^1$~~~~Liqiang Nie$^2$ \vspace{2mm}\\
$^1$Shandong University\hspace{1.5cm}$^2$Harbin Institute of Technology (Shenzhen)\hspace{1.5cm}\\
{\tt\small \{lizixu.cs,zivczw,fuzhiheng8,nieliqiang\}@gmail.com;} \\ 
{\tt\small mingyuzhang@mail.sdu.edu.cn,huyupeng@sdu.edu.cn}\\
{\tt\small \url{https://github.com/Lee-zixu/ConeSep/}}
}

\begin{document}
% \begin{CJK}{UTF8}{gbsn}

\twocolumn[{%
\maketitle
\begin{figure}[H]
\hsize=\textwidth % cvpr 需要
\centering
\includegraphics[width=\textwidth]{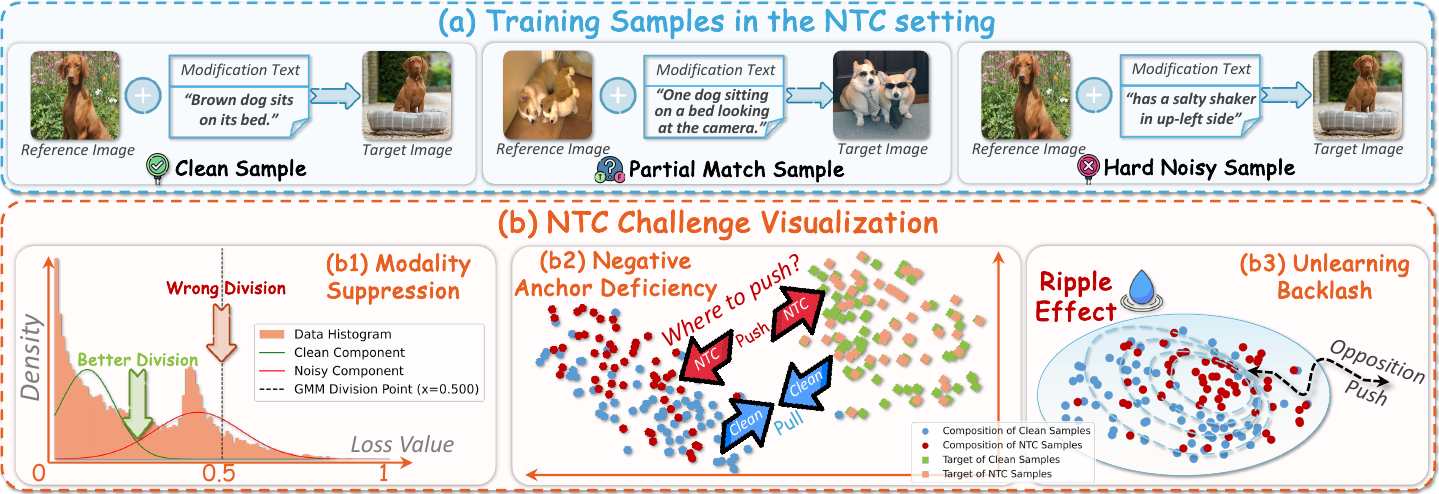}
\vspace{-20pt}
\caption{(a) illustrates examples of ``Clean Sample'', ``Partial Match Sample'' and ``Hard Noisy Sample'' within the NTC scenario. (b) illustrates the three challenges that current CIR methods face in the NTC scenario, including (b1) Modality Suppression, (b2) Negative Anchor Deficiency, and (b3) Unlearning Backlash.}
% \vspace{-7pt}
\label{fig:intro}
\end{figure}
}]

\blfootnote{$^*$ Corresponding Author: Yupeng Hu}

\begin{abstract}
The Composed Image Retrieval (CIR) task provides a flexible retrieval paradigm via a reference image and modification text, but it heavily relies on expensive and error-prone triplet annotations. This paper systematically investigates the Noisy Triplet Correspondence (NTC) problem introduced by annotations. We find that NTC noise, particularly ``hard noise'' (i.e., the reference and target images are highly similar but the modification text is incorrect), poses a unique challenge to existing Noise Correspondence Learning (NCL) methods because it breaks the traditional ``small loss hypothesis''. We identify and elucidate three key, yet overlooked, challenges in the NTC task, namely \textbf{(C1) Modality Suppression}, \textbf{(C2) Negative Anchor Deficiency}, and \textbf{(C3) Unlearning Backlash}. To address these challenges, we propose a \textbf{Cone}-based robu\textbf{S}t nois\textbf{E}-unlearning com\textbf{P}ositional network (ConeSep). Specifically, we first propose \textbf{Geometric Fidelity Quantization}, theoretically establishing and practically estimating a noise boundary to precisely locate noisy correspondence. Next, we introduce \textbf{Negative Boundary Learning}, which learns a ``diagonal negative combination'' for each query as its explicit semantic opposite-anchor in the embedding space. Finally, we design \textbf{Boundary-based Targeted Unlearning}, which models the noisy correction process as an optimal transport problem, elegantly avoiding Unlearning Backlash. Extensive experiments on benchmark datasets (FashionIQ and CIRR) demonstrate that ConeSep significantly outperforms current state-of-the-art methods, which fully demonstrates the effectiveness and robustness of our method.
\end{abstract}

\section{Introduction}
Composed Image Retrieval (CIR) allows users to retrieval target image \textit{(tar)} using a reference image \textit{(ref)} and a modification text \textit{(mod)}, thereby achieving flexible image retrieval~\cite{TME,tgcir,HINT,MELT,retrack} and gaining attention in computer vision~\cite{ERASE,song1,xiao2025visual,yu2026dinov3,xie2025chat,song2,tian2025core,cui2024correlation,jing2023category,jing2023multimodal,jiang2025self}, semantic understanding~\cite{wang2026fbs,wang2026tracking,dong2025aurora,song5,xie2026delving,zhang2026expseek,xiao2026not,ma2025mutuallearninghashingunlocking,Long_2026,lin2025se,song13,liu2025fusion} and multimodal learning~\cite{xie2026conquer,zhang2026towards,STABLE,song3,liu2024synthvlm,meng2026tri,song6,ma2026stableexplainablepersonalitytrait,li2025curriculum,xie2026hvd,song7,lin2026scientific,tian2025open,jiang2024prior,xu2025hdnet,yu2025iidm}. However, the performance of CIR highly relies on high-quality \textit{(ref, mod, tar)} triplet data. Obtaining such data is costly and challenging due to annotation expenses~\cite{FashionIQ, cirr}. Both the subjective bias from manual annotation and the hallucination effects introduced by using LVLMs~\cite{CASE, sprc,liu2025uniform,song4,liu2026chartverse,jiang2026foeforesterrorsmakes,wang2025ascd,fu2026maspo,song8,lin2026mmfinereason,bi2025cot,bi2025prismselfpruningintrinsicselection,chen2025autoneural,sun2023hierarchical,yu2025cotextor,10.1145/3746027.3755817} can lead to semantic inconsistency between the modification text \textit{(mod)} and the \textit{(ref, tar)} image pair. Li et al.~\cite{TME} recently define this problem systematically as Noisy Triplet Correspondence (NTC), which aims to train robust CIR models in noisy environments.

The NTC problem is intrinsically more complex than the widely studied Noisy Correspondence Learning (NCL) in other fields, particularly Noisy Dual Correspondence (NDC)~\cite{sun2024robust,yuan2025prototype,li2024incomplete,qin2023cross,li2023cross, npc}. Unlike NDC's simple pair mismatch (image-text or video-text), NTC features a composite noise structure as shown in Figure~\ref{fig:intro}(a), including partial matching (the \textit{mod} matches only the \textit{ref} or \textit{tar}) and hard noise (high \textit{ref/tar} visual similarity despite an incorrect \textit{mod}). This unique noise structure makes it difficult for traditional NDC methods, which rely on the ``small-loss hypothesis''~\cite{TME, npc}, to effectively separate the noise. Hard noisy samples, in particular, may exhibit a small loss value due to the strong visual similarity between \textit{ref} and \textit{tar}, causing them to be incorrectly classified as clean samples.

Existing NCL methods, including both the pioneering work in the NTC domain~\cite{TME} and mature techniques in the NDC domain~\cite{RCL, npc}, all demonstrate fundamental limitations when addressing NTC. Most of them follow the ``identification-correction/suppression'' paradigm. First, they rely on a coarse-grained mixed metric (such as a mixed loss or structural similarity) to partition clean and noisy data~\cite{TME, npc,song9}. Then, they either correct the noisy samples (e.g., by generating pseudo labels~\cite{NIC, TME}) or suppress them (e.g., by using robust loss functions~\cite{RCL, RDE}). However, we argue that this paradigm exposes three key, yet overlooked challenges within the NTC problem.

\textbf{C1: Modality Suppression.} In CIR, the query is composed of dense visual features (\textit{ref}) and sparse semantic instructions (\textit{mod}). In hard noise triplets, the strong visual similarity between (\textit{ref, tar}) tends to dominate the composed features, thereby suppressing the \textit{mod}'s weak mismatch signal. As shown in Figure~\ref{fig:intro}~(b1), this phenomenon leads to a \textit{Coarse-grained Metric}, making methods that rely on the mixed loss value to distinguish noise (e.g., TME's GMM classifier) highly susceptible to misclassifying this hard noise as a clean sample. Consequently, these samples escape filtering and compromise model training.

\textbf{C2: Negative Anchor Deficiency.} 
The difficulty in identifying ``hard noise'' in C1 exposes a more general problem: Even if some hard noise is successfully identified by chance, how should we handle this type of noise, which possesses a significant semantic gap? Common correction strategies attempt to \textit{pull back} the noisy modification towards some pseudo-ground-truth direction, but the effectiveness of such correction is questionable and may even lead to semantic confusion. We argue that a more reasonable strategy might be \textit{Targeted Unlearning}, which involves actively pushing the model away from the semantic direction of that noisy modification. However, this immediately introduces a new challenge: where should the model be pushed away to? As shown in Figure~\ref{fig:intro}~(b2), existing CIR frameworks (including TME) generally focus only on positive alignment, thus lacking a structured negative semantic anchor or boundary to push away from.

\textbf{C3: Unlearning Backlash.} Even if we resolve C2 by defining a negative anchor for hard noisy samples and attempt to push them away during training, we must recognize that the metric space is continuous and locally crowded.
As shown in Figure~\ref{fig:intro}(b3), this \textit{targeted unlearning} or \textit{push-away} operation may produce a ``ripple effect'', causing collateral damage to the representations of clean samples in the vicinity.
This presents a fundamental trade-off that must be overcome for robust NTC learning: How can one execute precise unlearning while simultaneously avoiding damage to ``healthy tissue''?

To systematically address the three aforementioned challenges, we require a feature space that enables fine-grained perception (for C1), structured repulsion (for C2), and avoids backlash (for C3). This requires us to not only define negative anchors but also to have a clear semantic decision boundary. Inspired by this, we propose a \textbf{Cone}-based robu\textbf{S}t nois\textbf{E}-unlearning com\textbf{P}ositional network (ConeSep), which consists of three logically progressive modules: (a) \textit{Geometric Fidelity Quantization (GFQ)}, which mitigates the impact of modality suppression (C1) on noise determination through noise boundary estimation; (b) \textit{Negative Boundary Learning (NBL)}, which resolves the negative anchor deficiency problem of C2 by learning a \textit{diagonal negative combination}; (c) \textit{Boundary-based Targeted Unlearning (BTU)}, which models the correction process as an optimal transport problem, elegantly circumventing the unlearning backlash problem of C3. Detailed descriptions are provided in \textit{Sec.~\ref{sec:method}, Methodology}.

In summary, our contributions include:
\begin{itemize}
    \item We identify and systematically articulate triple challenges of NTC problem in CIR, including modality suppression, negative anchor deficiency, and unlearning backlash.
    
    \item We propose ConeSep, a novel three-stage framework based on Cone-space Geometry, which provides an innovative solution for the NTC problem.

    \item Extensive experiments on two CIR benchmark datasets demonstrate that our proposed ConeSep significantly outperforms current SOTA methods under various noise ratios, demonstrating its superiority.
\end{itemize}

\begin{figure*}[ht!]
  \centering
     \vspace{-5pt}
   \includegraphics[width=\linewidth]{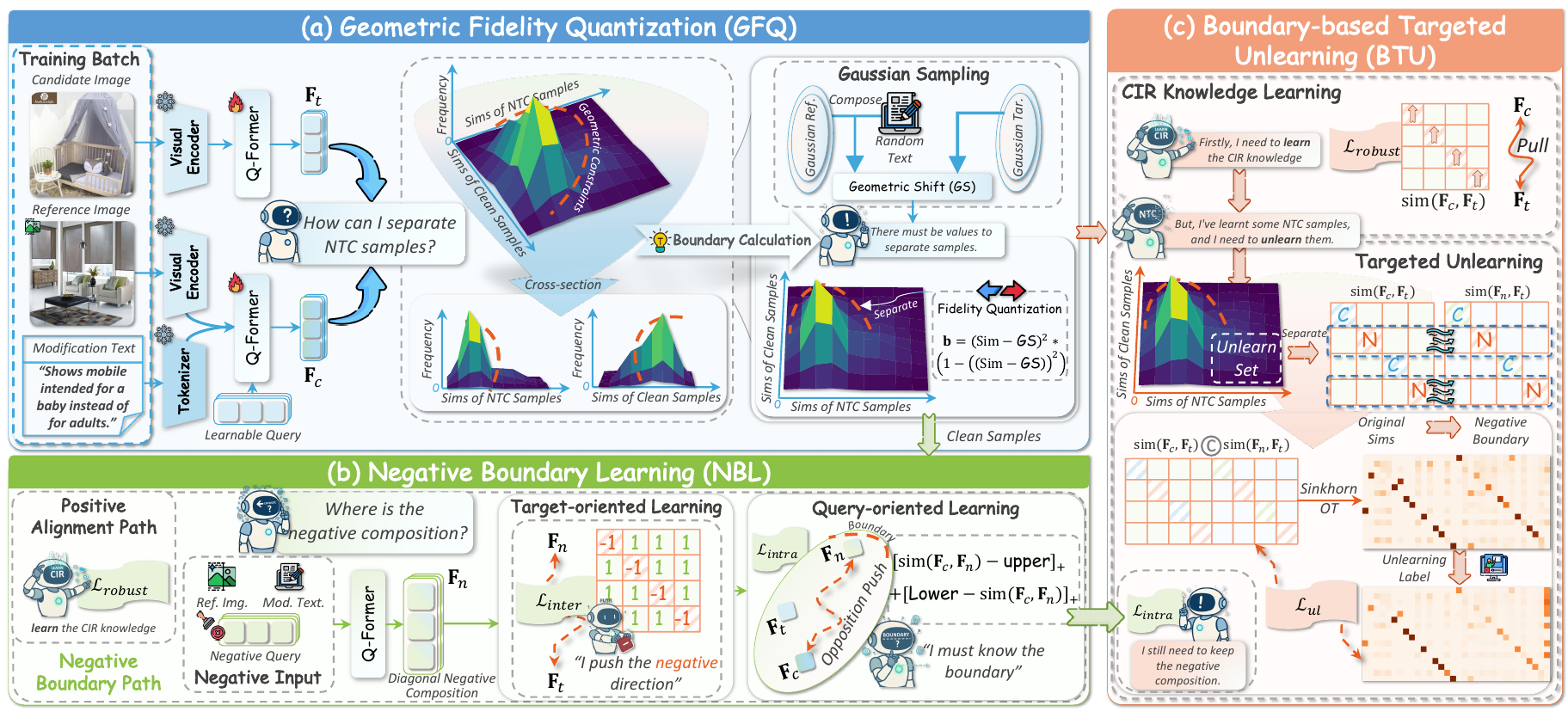}
      \vspace{-24pt}
   \caption{The proposed ConeSep consists of three primary modules: (a) Geometric Fidelity Quantization, (b) Negative Boundary Learning, and (c) Boundary-based Targeted Unlearning. In (a), we visualize the similarity cone effect, where the \textit{x} and \textit{y} axes denote the similarity scores of clean samples and NTC samples, respectively, and the \textit{z} axis shows the frequency distribution of these similarity values. The light green region in the figure forms a distinctly cone-shaped space.}
   \label{fig:framework}
   \vspace{-17pt}
\end{figure*}

\section{Related Work}

\statement{CIR with Noisy Correspondence.}
Composed Image Retrieval (CIR), valued for capturing complex user intentions~\cite{tgcir, sprc, TME}, has evolved from early fusion networks~\cite{HUD,REFINE,clvcnet} to modern Vision Language Pretraining (VLP) models like CLIP~\cite{clip}, BLIP~\cite{blip}, and BLIP-2~\cite{blip-2}. Research on CIR is expected to contribute to various applications, such as semantic understanding~\cite{li2024optimizing,cheng2026regime,zhong2026collaborativemultiagentscriptsgeneration,liu2024unsupervised,dong2026neureasonerexplainablecontrollableunified,li2024videocogqa,wang2026eeo,yu2025yielding}, multimodal retrieval~\cite{liu2020not,liu2021improving,xiao2025prompt,li2023ultrare,yuan2025video,zhang2026decoding,zhang2023multi,zhang2024cf,bi2025llava,wang2024twin,yu2025visualizing,10888444,li2025multi,lu2022understanding}.
However, most studies assume accurate triplet correspondence. In reality, large-scale datasets contain significant Noisy Triplet Correspondence (NTC) from collection and annotation errors. NTC is more complex than traditional noisy label problems~\cite{npc} and severely impacts robustness.
While recent work addresses NTC via sample selection or realignment~\cite{TME}, these methods only focus on discriminative robustness during new knowledge acquisition. They crucially fail to handle noisy samples that model has already learned, limiting performance.

\statement{Machine Unlearning.}
Machine Unlearning (MU) aims to modify a model to remove specific data's influence, emulating retraining from scratch~\cite{un1,zeng2025janusvln,zeng2025FSDrive,Xie_2025_ICCV}. Due to the high cost of exact unlearning, research focuses on efficient approximate methods~\cite{es,gsq}.
MU research falls into two categories. First, \textit{Unimodal Unlearning} targets specific image classes or data points~\cite{7, 9} or uses Gradient Ascent (GA) in LLMs to forget harmful knowledge~\cite{kga,yuan2026if,yuan2025autodrive,jiang2025transforming,lu2024generic}. These methods forcibly push away the feature representations of targeted samples. Second, \textit{Multimodal Unlearning} is a more nascent field~\cite{26, 27}, addressing cross-modal categories or generative models~\cite{forgetme}, but is often limited by the complexities of modality heterogeneity.
A key issue is that current methods, such as Gradient Ascent, rely on local, non-targeted ``push away'' operations. This vulnerability leads to ``Unlearning BackLash,'' where correcting one error damages knowledge of surrounding clean samples. In contrast, our proposed ConeSep avoids this blind ``push away.'' It uses optimal transport to find a smooth, optimal path, shifting the probability distribution from a noisy sample to a newly defined inverse semantic boundary. This achieves targeted unlearning of hard noise without drastic feature space perturbations, thereby fundamentally circumventing unlearning backlash.

\section{Methodology}
\label{sec:method}

As a core innovation, our proposed ConeSep is designed to construct a closed-loop system capable of actively perceiving, structurally modeling, and precisely ``unlearning'' noise.
As shown in Figure~\ref{fig:framework}, ConeSep consists of three logically modules, namely: 
(a) \textit{Geometric Fidelity Quantization} (GFQ, detailed in Section~\ref{subsubsec:Geometric Fidelity Quantization}), (b) \textit{Negative Boundary Learning} (NBL, detailed in Section~\ref{subsubsec:Negative Boundary Learning}), and (c) \textit{Boundary-based Targeted Unlearning } (BTU, detailed in Section~\ref{subsubsec:Boundary-based Targeted Unlearning}).
In this section, we first provide the problem definition and them elaborate on each module of ConeSep.

\subsection{Problem Formulation}
Composed Image Retrieval (CIR) is a task designed to retrieve the target image that matches a multimodal query~\cite{pair,median,encoder,OFFSET,FineCIR}. However, CIR datasets in practical applications often contain annotation noise, a situation referred to as \textit{Noisy Triplet Correspondence (NTC)}.
We follow the experimental setup of TME~\cite{TME} and simulate noisy scenarios by sampling subsets based on a noise rate $\sigma$ and randomly shuffling the samples. Given a set of triplets with NTC, $\mathcal{T} = \{\langle x_r, x_m, x_t \rangle_n\}_{n=1}^N$, our goal is to learn a robust embedding function $\mathcal{G}$, which maps the multimodal query $(x_r, x_m)$ and its corresponding target image $x_t$ into a shared metric space such, denoted as $\mathcal{G}(x_r, x_m) \rightarrow \mathcal{G}(x_t)$.

\subsection{Geometric Fidelity Quantization (GFQ)} 
\label{subsubsec:Geometric Fidelity Quantization}
To address the challenge of modality suppression, we design the \textit{Geometric Fidelity Quantization (GFQ)} module. This module is based on the theory of noise boundary and utilizes the geometric separability of the cone space to explicitly localize the geometric noise boundary, thereby quantifying the matching fidelity of each sample.

\statement{Feature Extraction.}
Following previous works~\cite{sprc, TME}, we utilize the Q-Former from BLIP-2~\cite{blip-2} to extract the multimodal composed feature $\mathbf{F}_c$ and the target image feature $\mathbf{F}_t$, formalized as follows,
\begin{equation}
\small
\mathbf{F}_c\!=\!\mathcal{P}_c(\text{Q-F}(\Phi_{\mathbb{I}}(x_r),\!\Phi_{\mathbb{T}}(x_m))),
    \mathbf{F}_t\!=\!\mathcal{P}_t(\text{Q-F}(\Phi_{\mathbb{I}}(x_t))),
\label{eq:features}
\end{equation}
where $\mathbf{F}_c, \mathbf{F}_t\in \mathbb{R}^{Q \times D}$ are the multimodal composed feature and target feature, respectively. $\text{Q-F}$ is the Q-Former, $Q$ is the number of learnable queries in the Q-Former, and $D$ is the embedding dimension. $\Phi_{\mathbb{I}}$ and $\Phi_{\mathbb{T}}$ are the image encoder and text tokenizer, respectively, and $\mathcal{P}_c$ and $\mathcal{P}_t$ are the composed query mapping layer and the target image mapping layer, respectively. $x_r$, $x_m$, and $x_t$ are the reference image, modification text, and the corresponding target image, respectively.

\statement{Fidelity Quantization.}
To mitigate the issue of inaccurate noise judgment caused by modality suppression, we first identify a boundary that can accurately distinguish noise samples and then quantize the fidelity of the training samples.
Specifically, we employ a random sampling method to estimate the similarity boundary $\mathbb{B}$. Specifically, we perform $K$ Gaussian samplings $x_{r,k}^{G}, x_{t,k}^{G} \sim \mathcal{N}(0,1)$ for the reference image and the target image, and randomly sample the modification text $x_{m,k}^{G}$ within the batch. Then, we perform feature extraction on these samples via Eq.(\ref{eq:features}) and calculate their average cosine similarity as the estimation of boundary~$\mathbb{B}$, formulated as,
\begin{equation}
\small
\mathbb{B}\!\!=\!\!\frac{\sum_{j=1}^{K} s(\mathcal{P}_c(\text{Q-F}(\Phi_{\mathbb{I}}(x_{r,k}^{G}),\!\Phi_{\mathbb{T}}(x_{m,k}^{G}))), \mathcal{P}_t(\text{Q-F}(\Phi_{\mathbb{I}}(x_{t,k}^{G}))))}{K},
\end{equation}
where $s(\cdot)$ is the cosine similarity, and the remaining symbols share the same definitions as in Eq.(\ref{eq:features}). Based on the boundary $\mathbb{B}$, we define the following fidelity measure function $\mathcal{F}(\textbf{F}_c, \textbf{F}_t)$, which quantifies the fidelity of a sample being a matching sample, formulated as,
\begin{equation}
% \small
    \mathcal{F}(\textbf{F}_c,\!\textbf{F}_t)\!\!=\!\!(\text{ReLU}(s_{ct}\!-\!\mathbb{B}))^2\!\cdot\! (\text{ReLU}(s_{ct}-\!\mathbb{B})\!-\!1),
\label{eq:fidelity}
\end{equation}
where $\text{ReLU}(\cdot)\!\!=\!\![\cdot]_+$ and $s_{ct}\!\!=\!\!\cos(\textbf{F}_c,\textbf{F}_t)$. A larger value of $\mathcal{F}(\textbf{F}_c, \textbf{F}_t)$ indicates that the sample is more likely to be a clean sample. Finally, we set a fidelity threshold $\omega$ to divide the batch samples into a high fidelity clean set $\mathcal{T}_{clean}= \{\langle x_r,x_m,x_t \rangle_b | \mathcal{F}(\textbf{F}_{c,b}, \textbf{F}_{t,b}) \ge \omega\}$ and a low fidelity noise set $\mathcal{T}_{noisy}=\{\langle x_r,x_m,x_t \rangle_b | \mathcal{F}(\textbf{F}_{c,b}, \textbf{F}_{t,b}) < \omega\}$.

\subsection{Negative Boundary Learning (NBL)}
\label{subsubsec:Negative Boundary Learning}
To address the challenge of missing negative anchors, we introduce the \textit{Negative Boundary Learning (NBL)} module. Specifically, we propose a dual path learning strategy: \textit{Positive Alignment Path} is designed to the standard CIR paradigm, and the \textit{Negative Boundary Learning Path} is dedicated to constructing structured negative boundaries. The design of these two paths is presented below.

\statement{Positive Alignment Path.}
First, we design this path to ensure that the model consistently learns the core paradigm of CIR, which is promoting the composed feature $\textbf{F}_c$ closer to its corresponding target image feature $\textbf{F}_t$ for clean samples. Inspired by RCL~\cite{RCL}, we adopt the \textit{Robust Contrastive Loss} to reduce the influence of negative samples during the alignment process, thereby robustly pushing the composed feature closer to the target feature, formulated as,
\begin{equation}
\small
\mathcal{L}_{robust} = 
- \frac{1}{B} \sum_{i,j\neq i}^{B} \log
(1-\frac{
    \exp\big( s(\textbf{F}_{ci}, \textbf{F}_{tj} / \tau \big)
}{
    \sum_{j=1}^{B} \exp\big(  s(\textbf{F}_{ci}, \textbf{F}_{tj} / \tau \big)
}),
\label{robust}
\end{equation}
where $\tau$ is the temperature coefficient, $B$ is the batch size, $s(\cdot)$ is the cosine similarity function. $\mathbf{F}_c^i$ and $\mathbf{F}_t^i$ are the $i$-th composed feature and target image feature, respectively.

\statement{Negative Boundary Path.}
Then, we construct the this path to build a \textit{Diagonal Negative Composition} $\mathbf{F}_{neg}$ for each composed query, which represents the semantic opposite of the query in the metric space, to serve as an orientation guide for pushing away noise boundaries. Specifically, we introduce a set of learnable $\text{negative prompt}$ $\mathbf{P}_{\text{neg}} \in \mathbb{R}^{Q \times D}$ and, similar to Eq.($\ref{eq:features}$), formulate the negative composed feature $\mathbf{F}_{neg}$ \textit{i.e., Diagonal Negative Composition}, as follows,
\begin{equation}
\mathbf{F}_{neg}\!=\!\mathcal{P}_c(\text{Q-F}(\mathbf{P}_{\text{neg}}, \Phi_{\mathbb{I}}(x_r),\!\Phi_{\mathbb{T}}(x_m))).
\end{equation}

Subsequently, we optimize the construction of the \textit{Diagonal Negative Composition} by focusing on moving it away from both the target feature and the composed feature.

\textit{Target-oriented Learning.}
This aims to shape $\mathbf{F}_{neg}$ as the ``opposite'' of $\mathbf{F}_t$. Inspired by reverse matching strategy of Sigmoid Loss~\cite{sigmoid_loss}, we define a binary target matrix $\mathbf{T}$, where $\mathbf{T}_{ij} = 1$ (when $i=j$) and $\mathbf{T}_{ij} = -1$ (when $i \neq j$). The objective is that $s(\mathbf{F}_{neg}^i, \mathbf{F}_t^j)$ approximates the inverse target $-\mathbf{T}_{ij}$. This strategy ensures that \textit{Diagonal Negative Composition} $\mathbf{F}_{neg}$ moves \textit{away from} its corresponding target ($i=j$) and \textit{close to} all non-matching target images ($i \neq j$). Thus, it becomes the opposite of corresponding query's semantics in metric space, formulated as follows,
\begin{equation}
\small
\mathcal{L}_{inter}\!\!=\!\!\frac{\sum_{i \in \mathcal{T}_{clean}}\!\!\sum_{j=1}^B \log(1\!\!+\!\!e^{(\textbf{T}_{ij} \cdot s(\mathbf{F}_{neg}^i, \mathbf{F}_t^j) / \tau)})}{|\mathcal{T}_{clean}| \cdot B}.
\end{equation}

\textit{Query-oriented Learning.} 
Furthermore, the \textit{Diagonal Negative Composition} $\mathbf{F}_{neg}$ should also be semantically opposite to corresponding original composed feature $\mathbf{F}_{c}$. We employ a relaxed similarity boundary that constrains similarity $s(\mathbf{F}_{c}, \mathbf{F}_{neg})$ between $\mathbf{F}_{c}$ and $\mathbf{F}_{neg}$ to fall within a preset interval $[\alpha_1, \alpha_2]$ (where $\alpha_1$ is mean of similarities less than zero within batch samples, and $\alpha_2$ is the mean of similarities greater than zero within batch samples). This causes similarity to hover around $0$, achieving an effect where two vectors are orthogonally distant, formulated as follows,
\begin{equation}
\small
\mathcal{L}_{intra}\!\!=\!\!\frac{\sum_{i\in \mathcal{T}_{clean}}\![\text{ReLU}( s_n + \alpha_1)\!\!+\!\!\text{ReLU}(-\alpha_2 - s_n)]}{|\mathcal{T}_{clean}|},
\label{intra}
\end{equation}
where $s_n=\cos(\mathbf{F}_c^i, \mathbf{F}_{neg}^i)$.
Finally, the objective function for the NBL module is obtained as follows,
\begin{equation}
\Theta^* = \arg\min_{\Theta} ( \mathcal{L}_{rank} + \zeta \mathcal{L}_{intra} + \nu \mathcal{L}_{inter}
),
\end{equation}
where $\Theta^*$ represents the parameters of the NBL module being optimized, and $\zeta$ and $\nu$ are hyperparameters. It is noteworthy that the objective function presented above is utilized only for $N$ epochs during the initial warm-up phase.

\begin{table*}[ht]
\centering
\tabcolsep=10pt
\caption{Performance comparison on the FashionIQ validation set in terms of R@K(\%). The best and sub-optimal results are highlighted in \textbf{bold} and \underline{underline}, respectively.}
\vspace{-8pt}
\resizebox{0.88\linewidth}{!}{%
\begin{tabular}{c|l|cc|cc|cc|cc|ccc}
\Xhline{1.5pt}
\multirow{2}{*}{\textbf{Noise}} & \multirow{2}{*}{\textbf{Methods}}
& \multicolumn{2}{c|}{\textbf{Dress}}
& \multicolumn{2}{c|}{\textbf{Shirt}}
& \multicolumn{2}{c|}{\textbf{Toptee}}
& \multicolumn{3}{c}{\textbf{Average}} \\
\cline{3-11}
& & R@10 & R@50 & R@10 & R@50 & R@10 & R@50 & R@10 & R@50 & AVG. \\
\hline
\hline

& SSN~\cite{ssn}~(AAAI'24) & 34.36 & 60.78 & 38.13 & 61.83 & 44.26 & 69.05 & 38.92 & 63.89 & 51.40 \\

& CALA~\cite{cala}~(SIGIR'24) & 42.38 & 66.08 & 46.76 & 68.16 & 50.93 & 73.42 & 46.69 & 69.22 & 57.96 \\

& SPRC~\cite{sprc}~(ICLR'24) & 49.18 & \underline{72.43} & 55.64 & 73.89 & \underline{59.35} & 78.58 & 54.92 & 74.97 & 64.85 \\

0\% & TME~\cite{TME}~(CVPR'25) & {49.73} & 71.69 & {56.43} & {74.44} & {59.31} & \underline{78.94} & {55.15} & {75.02} & {65.09} \\

& {HABIT~\cite{habit} (AAAI'26)} & {49.99} & 72.38 & \underline{56.62} & {74.68} & \textbf{59.51} & 78.53 & \underline{55.38} & \underline{75.20} & \underline{65.29}\\

& {INTENT~\cite{intent} (AAAI'26)} & \underline{50.32} & {72.10} & {56.32} & \underline{74.93} & {59.28} & {78.45} & {55.31} & {75.16} & {65.24} \\
% \cline{2-11}

& \cellcolor{noise80color}\textbf{ConeSep (Ours)} & \cellcolor{noise80color}\textbf{50.96} & \cellcolor{noise80color}\textbf{73.02} & \cellcolor{noise80color}\textbf{56.98} & \cellcolor{noise80color}\textbf{75.22} & \cellcolor{noise80color} 58.80 & \cellcolor{noise80color}\textbf{79.40} & \cellcolor{noise80color}\textbf{55.58} & \cellcolor{noise80color}\textbf{75.88} & \cellcolor{noise80color}\textbf{65.73}\\
\hline

% --- 20% 噪声块 (浅绿色) ---
& SSN~\cite{ssn}~(AAAI'24) & 22.61 & 45.56 & 27.87 & 48.58 & 31.82 & 55.28 & 27.43 & 49.81 & 38.62 \\
& CALA~\cite{cala}~(SIGIR'24) & 29.05 & 51.36 & 35.28 & 56.23 & 36.05 & 58.24 & 33.46 & 55.28 & 44.37 \\
& SPRC~\cite{sprc}~(ICLR'24) & 39.81 & 62.22 & 48.58 & 66.29 & 50.48 & 70.58 & 46.29 & 66.36 & 56.33 \\
20\% & TME~\cite{TME}~(CVPR'25)& {49.03} & 70.35 & \underline{55.84} & {73.16} & {57.22} & {78.23} & {54.03} & {73.91} & {63.97} \\

& HABIT~\cite{habit} (AAAI'26) & \underline{49.63}& {71.34} & {55.67}& {73.19 }& \underline{58.14} & {78.32} & \underline{54.48} & {74.28} & \underline{64.38} \\

& {INTENT~\cite{intent} (AAAI'26)} & {49.32}& \underline{71.43} & {55.32}& \underline{73.57}& {58.01} & \textbf{78.46} & {54.22} & \underline{74.49} & {64.36} \\
& \cellcolor{noise80color}\textbf{\modelname (Ours)} & \cellcolor{noise80color}\textbf{50.23}& \cellcolor{noise80color}\textbf{72.19} & \cellcolor{noise80color}\textbf{56.28}& \cellcolor{noise80color}\textbf{74.45}& \cellcolor{noise80color}\textbf{58.29} & \cellcolor{noise80color}\underline{78.38} & \cellcolor{noise80color}\textbf{54.93} & \cellcolor{noise80color}\textbf{75.01} & \cellcolor{noise80color}\textbf{64.97} \\
\hline
% --- 50% 噪声块 (浅橙色) ---
 & SSN~\cite{ssn}~(AAAI'24) & 15.27 & 33.71 & 23.36 & 41.61 & 22.79 & 42.94 & 20.47 & 39.42 & 29.95 \\
 & CALA~\cite{cala}~(SIGIR'24) & 20.77 & 40.95 & 29.69 & 46.57 & 27.03 & 46.81 & 24.83 & 44.78 & 34.80 \\
 & SPRC~\cite{sprc}~(ICLR'24) & 35.94 & 57.16 & 42.25 & 61.63 & 44.98 & 64.76 & 41.06 & 61.19 & 51.12 \\
  50\% & TME~\cite{TME}~(CVPR'25)& {46.26} & {68.27} & {53.09} & {71.88} & {55.07} & {76.59} & {51.47} & {72.25} & {61.86} \\

& HABIT~\cite{habit} (AAAI'26) & {47.33} & {69.71} & \underline{53.72} & \underline{72.55} & {56.51} & \underline{77.00} & \underline{52.52} & {73.09} & {62.80} \\

& \textbf{INTENT~\cite{intent} (AAAI'26)} & \underline{47.99} & \underline{71.24} & {52.78} & {72.48} & \underline{56.79} & {76.23} & \underline{52.52} & \underline{73.32} & \underline{62.92} \\
 & \cellcolor{noise80color}\textbf{\modelname (Ours)} & \cellcolor{noise80color}\textbf{48.39} & \cellcolor{noise80color}\textbf{71.34} & \cellcolor{noise80color}\textbf{54.56} & \cellcolor{noise80color}\textbf{73.25} & \cellcolor{noise80color}\textbf{57.22} & \cellcolor{noise80color}\textbf{77.87} & \cellcolor{noise80color}\textbf{53.39} & \cellcolor{noise80color}\textbf{74.15} & \cellcolor{noise80color}\textbf{63.77} \\
\hline
% --- 80% 噪声块 (浅红色) ---
 & SSN~\cite{ssn}~(AAAI'24) & 11.16 & 25.24 & 16.98 & 30.72 & 17.03 & 32.64 & 15.05 & 29.53 & 22.29 \\
 & CALA~\cite{cala}~(SIGIR'24) & 14.28 & 30.59 & 19.73 & 35.82 & 19.48 & 36.10 & 17.83 & 34.41 & 26.00 \\
 & SPRC~\cite{sprc}~(ICLR'24) & 28.41 & 50.77 & 36.21 & 54.37 & 35.90 & 59.06 & 33.51 & 54.03 & 43.77 \\
 80\% & TME~\cite{TME}~(CVPR'25)& {41.45} & {64.35} & {47.30} & {68.20} & {51.25} & {73.23} & {46.67} & {68.60} & {57.63} \\

& HABIT~\cite{habit}(AAAI'26) & {42.04} & {65.20} & {50.12} & \underline{69.77} & {52.92} & {73.61} & {48.36} & {69.53} & {58.94} \\

& {INTENT~\cite{intent} (AAAI'26)} & \underline{42.07} & \underline{65.58} & \underline{50.38} & {69.41} & \underline{53.09} & \underline{73.91} & \underline{48.51} & \underline{69.63} & \underline{59.07} \\
 & \cellcolor{noise80color}\textbf{\modelname (Ours)} & \cellcolor{noise80color}\textbf{43.33} & \cellcolor{noise80color}\textbf{66.01} & \cellcolor{noise80color}\textbf{51.23} & \cellcolor{noise80color}\textbf{70.35} & \cellcolor{noise80color}\textbf{53.19} & \cellcolor{noise80color}\textbf{74.55} & \cellcolor{noise80color}\textbf{49.25} & \cellcolor{noise80color}\textbf{70.30} & \cellcolor{noise80color}\textbf{59.77} \\
\Xhline{1.5pt}
\end{tabular}
}
\vspace{-15pt}
\label{tab:fiq_noise}
\end{table*}

\subsection{Boundary-based Targeted Unlearning (BTU)}
\label{subsubsec:Boundary-based Targeted Unlearning}
To avoid the ``ripple effect'' generated during the process of pushing away noise from harming the clean sample features in the space, we design the \textit{Boundary-based Targeted Unlearning (BTU)} module. This module utilizes the diagonal negative composition $F_{neg}$ learned by \textit{NBL} as an anchor, and it performs precise targeted unlearning through Optimal Transport (OT) to prevent the ``unlearning backlash''.

Specifically, we utilize the low-fidelity noise set $\mathcal{T}_{noisy}$ obtained from the \textit{GFQ} module as the objective for targeted unlearning. Simultaneously, we model the computation of the rationally forgotten label distribution as an $B \times 2B$ Optimal Transport problem. For this purpose, we construct the joint cost matrix $\mathbf{C} = [\mathbf{C}^{+} | \mathbf{C}^{-}]$, which considers the cost from any in-batch sample $\mathbf{F}_c^i$ to all ``positive targets'' (i.e., target image $\mathbf{F}_t^j$) and ``negative boundaries'' (i.e., diagonal negative composition $\mathbf{F}_{neg}^j$), where the costs $\mathbf{C}_{ij}^{+}$ and $\mathbf{C}_{ij}^{-}$ are formulated as follows,
\begin{equation}
\textbf{C}_{ij}^{+} = 1 - s(\mathbf{F}_c^i, \mathbf{F}_t^j), \textbf{C}_{ij}^{-} = 1 - s(\mathbf{F}_c^i, \mathbf{F}_{neg}^j).
\end{equation}

To avoid the unlearning process from harming neighboring clean samples, we use the mask matrix $\mathbf{M}\in \mathbb{R}^{B \times 2B}$ to precisely \textit{sever} specific transport paths. \textbf{(1)} For low-fidelity noise samples $\mathcal{T}_{noisy}$, we block their path flowing to the corresponding positive target ($j\!\!=\!\!i$); \textbf{(2) }For high-fidelity clean samples $\mathcal{T}_{clean}$, we block their path flowing to the corresponding negative boundary ($j\!\!=\!\!i\!\!+\!\!B$) to protect their positive alignment. Formally, we set the mask value $\mathbf{M}_{ij}$ for a \textit{blocked} path to $1$ and the value for an \textit{unblocked} path to $0$.

This mask $\mathbf{M}$ is subsequently used to generate the final targeted unlearning cost matrix $\mathbf{C}_{masked}$. We achieve the \textit{severing} of blocked paths ($\mathbf{M}_{ij}=1$) by applying an extremely large cost ($\infty$), specifically defined as $[\mathbf{C}_{masked}]_{ij} = [\mathbf{C}]_{ij} + \mathbf{M}_{ij} \cdot \infty$. Based on this, our objective is to solve for an optimal transport plan $\mathbf{P} \in \mathbb{R}^{B \times 2B}$ that minimizes the overall \textit{targeted} transport cost from the composed feature $\mathbf{F}_c$ to the target image feature $\mathbf{F}_t$ and the negative boundary $\mathbf{F}_{neg}$. Formally, this entropy-regularized OT objective is defined as,
\begin{equation}
   \min_{\mathbf{P} \in \Pi(\mathbf{u}, \mathbf{v})} s(\mathbf{P}, \mathbf{C}_{masked})  - \epsilon H(\mathbf{P}),
\end{equation}
where $s(\cdot)$ is the cosine similarity function, $\mathbf{P}$ is the transport plan, $H(\mathbf{P})$ is the entropy of $\mathbf{P}$, $\mathbf{C}_{masked}$ is the masked $B \times 2B$ cost matrix, and $\epsilon$ is the entropy regularization parameter. $\Pi(\mathbf{u}, \mathbf{v})$ represents the set of all valid transport plans that satisfy the source marginal distribution $\mathbf{u} \in \mathbb{R}^B$ and the target marginal distribution $\mathbf{v} \in \mathbb{R}^{2B}$. This objective function is quickly solvable through the efficient Sinkhorn-Knopp algorithm~\cite{Sinkhorn-Knopp}, iteratively finding the optimal transport plan $\mathbf{P}^*$.

Subsequently, to preserve a strong supervisory signal and mitigate neighbor sample damage, we construct a smooth soft label $\mathbf{Y}$ which collectively targets forgetting and forward learning. The $\mathbf{Y}$ integrates the global optimal plan $\mathbf{P}^*$ from optimal transport (OT) and the original hard label $\mathbf{L} \in \mathbb{R}^{B \times 2B}$. The original hard label $\mathbf{L}$ contains ones on the diagonal, but for a row $i$ in the set of noisy samples $\mathcal{T}_{\text{noisy}}$, $\textbf{L}_{i, i}$ is set to zero and $\textbf{L}_{i, i+B}$ is set to one. The smooth soft label $\mathbf{Y}$ is formulated as follows,
\begin{equation}
\mathbf{Y} = \gamma \cdot \mathbf{P} + (1 - \gamma) \cdot \mathbf{L},
\end{equation}
where $\gamma$ is the balancing hyperparameter. Finally, we employ the Kullback-Leibler (KL) divergence to define the targeted forgetting loss $\mathcal{L}_{ul}$, which is formulated as follows,
\begin{equation}
\mathcal{L}_{ul} =\text{KL}(\log [s(\textbf{F}_c,\textbf{F}_t), s(\textbf{F}_{neg},\textbf{F}_t)] || \mathbf{Y}).
\end{equation}

Ultimately, we still require the model to learn the paradigm knowledge of CIR while maintaining the learning of negative composition. Thus, we integrate $\mathcal{L}_{\text{robust}}$~(Eq.(\ref{robust})) and $\mathcal{L}_{\text{intra}}$~(Eq.(\ref{intra})) to obtain the final optimization objective of ConeSep as follows,
\begin{equation}
\Psi^*= \arg\min_{\Psi_{}} \left( \mathcal{L}_{robust} + \kappa \mathcal{L}_{ul} + \zeta \mathcal{L}_{intra} \right),
\end{equation}
where $\Psi^*$ is the optimization parameters of ConeSep, and $\kappa$ and $\zeta$ are hyperparameters.

\section{Experiments}

This section presents the experiments conducted to evaluate the \modelname and the corresponding analyses. Following the previous work~\cite{TME,intent,airknow}, all ablation studies and parameter sensitivity analyses are conducted under a noise rate of $\sigma=0.2$.
%%%%%%%%%%%%%%%%%%%%%%%%%%%%%%%%%%%%%%%%%%%%%%%%%%%%%%%%%%%%%%%%%%
\subsection{Experimental Settings}

\statement{Datasets.} Our experimental evaluation is conducted on two CIR benchmarks, including fashion-domain dataset FashionIQ~\cite{FashionIQ} and open-domain dataset CIRR~\cite{cirr}.

\statement{Evaluation Metrics.} To evaluate the retrieval performance, we employ the widely-used Recall@K (R@K) as the core evaluation standard. For FashionIQ, we follow its standard metrics~\cite{FashionIQ}, reporting R@10 and R@50 for each of the three categories (\textit{Dresses, Shirts, Tops\&Tees}). For CIRR, we adopt R@$k$ ($k$=$1, 5, 10, 50$) and R$_{sub}$@$k$ ($k$=$1, 2, 3$).

\statement{Implementation Details.} Following previous works~\cite{TME,sprc}, our \modelname uses the BLIP-2~\cite{blip-2} model as its base architecture. \modelname is trained using the AdamW optimizer with an initial learning rate of $1e-5$ for CIRR and $2e-5$ for FashionIQ, and a batch size of $128$. The number of random samples $K$ = $4$ and the temperature coefficient $\tau$ = $0.07$. The fidelity threshold $\omega$=$0.5$, balancing parameter $\gamma=0.7$, and loss hyperparameters $\{\zeta,\nu,\kappa\}\!\!=\!\!0.5$. All experiments are conducted on one NVIDIA A40 GPU for $20$ epochs.
%%%%%%%%%%%%%%%%%%%%%%%%%%%%%%%%%%%%%%%%%%%%%%%%%%%%%%%%%%%%%%%%%%

\begin{table*}[t]
\centering
\tabcolsep=10pt
\caption{Performance comparison on the CIRR test set in terms of R@K(\%) and R$_{sub}$@K(\%). The best and second-best results are highlighted in \textbf{bold} and \underline{underline}, respectively.}
\vspace{-8pt}
\resizebox{0.9\linewidth}{!}{%
\begin{tabular}{c|l|cccc|ccc|c}
\Xhline{1.5pt}

\multirow{2}{*}{\textbf{Noise}} & \multirow{2}{*}{\textbf{Methods}}
& \multicolumn{4}{c|}{\textbf{R@K}}
& \multicolumn{3}{c|}{\textbf{R$_{sub}$@K}}
& \multirow{2}{*}{\textbf{Avg(R@5, R$_{sub}$@1)}} \\
\cline{3-9}
& & K=1 & K=5 & K=10 & K=50 & K=1 & K=2 & K=3 & \\
\hline
\hline

& SSN~\cite{ssn}~(AAAI'24) & 43.91 & 77.25 & 86.48 & 97.45 & 71.76 & 88.63 & 95.54 & 74.51 \\

& CALA~\cite{cala}~(SIGIR'24) & 49.11 & 81.21 & 89.59 & 98.00 & 76.27 & 91.04 & 96.46 & 78.74 \\

& SPRC~\cite{sprc}~(ICLR'24) & 51.96 & 82.12 & 89.74 & 97.69 & 80.65 & 92.31 & 96.60 & 81.39 \\

0\% & TME~\cite{TME}~(CVPR'25) & \textbf{53.42} & {82.99} & 90.24 & 98.15 & \underline{81.04} & {92.58} & {96.94} & \underline{82.01} \\
% \cline{2-10}

& HABIT~\cite{habit} (AAAI'26) & {52.71} & {82.64} & \underline{90.63} & {98.19} & {80.99} & \textbf{92.77} & \underline{97.00} & {81.82} \\

& {INTENT~\cite{intent} (AAAI'26)} & \underline{53.37} & \underline{83.16} & \textbf{90.73} & \underline{98.22} & {80.24} & {92.37} & {96.89} & {81.70} \\

& \cellcolor{noise80color}\textbf{ConeSep (Ours)} &  \cellcolor{noise80color}53.06 & \cellcolor{noise80color}\textbf{83.45} & \cellcolor{noise80color} {90.60} & \cellcolor{noise80color}\textbf{98.34} & \cellcolor{noise80color}\textbf{81.23} & \cellcolor{noise80color}\underline{92.62} & \cellcolor{noise80color}\textbf{97.24} & \cellcolor{noise80color}\textbf{82.34} \\
\hline

 & SSN~\cite{ssn}~(AAAI'24) & 34.02 & 65.90 & 75.78 & 91.33 & 66.92 & 85.90 & 93.45 & 66.41 \\
 & CALA~\cite{cala}~(SIGIR'24) & 41.33 & 72.70 & 82.84 & 94.34 & 71.66 & 88.15 & 94.94 & 72.18 \\
 & SPRC~\cite{sprc}~(ICLR'24) & 45.90 & 75.86 & 83.52 & 93.37 & 78.10 & {91.40} & 96.05 & 76.98 \\

 20\%  & TME~\cite{TME}~(CVPR'25) & {51.35} & 81.01 & 88.53 & {97.81} & \underline{78.46} & 91.25 & {96.39} & \underline{79.74} \\

& HABIT~\cite{habit} (AAAI'26) & \underline{51.68} & {81.02} & {89.24} & {97.81} & {78.20} & \underline{91.66} & \textbf{96.75} & {79.61} \\

& {INTENT~\cite{intent} (AAAI'26)} & {51.25} & \underline{81.36} & \textbf{90.02} & \underline{98.05} & {77.95} & {91.40} & \underline{96.46} & {79.66} \\
 & \cellcolor{noise80color}\textbf{\modelname (Ours)} & \cellcolor{noise80color}\textbf{52.29} & \cellcolor{noise80color}\textbf{82.19} & \cellcolor{noise80color}\underline{89.98} & \cellcolor{noise80color}\textbf{98.19} & \cellcolor{noise80color}\textbf{78.66} & \cellcolor{noise80color}\textbf{91.76} & \cellcolor{noise80color}\textbf{96.75} & \cellcolor{noise80color}\textbf{80.43} \\
\hline
% --- 50% 噪声块 (浅橙色) ---

 & SSN~\cite{ssn}~(AAAI'24) & 25.93 & 53.71 & 63.40 & 82.10 & 62.10 & 82.27 & 91.57 & 57.90 \\
 & CALA~\cite{cala}~(SIGIR'24) & 36.10 & 66.12 & 77.76 & 92.10 & 68.12 & 85.66 & 93.59 & 67.12 \\
 & SPRC~\cite{sprc}~(ICLR'24) & 39.93 & 66.00 & 73.59 & 86.48 & 75.81 & 89.21 & 95.37 & 70.90 \\

 50\% & TME~\cite{TME}~(CVPR'25) & 48.48 & {78.94} & {87.28} & {96.99} & {76.48} & {90.07} & {95.83} & {77.71} \\

& HABIT~\cite{habit} (AAAI'26) & \underline{50.32} & {79.63} & {88.34} & {97.06} & {76.84} & \underline{90.60} & \underline{96.27} & \textbf{78.87} \\

& INTENT~\cite{intent} (AAAI'26) & {49.78} & \underline{79.64} & \underline{88.99} & \underline{97.37} & \textbf{77.18} & {90.41} & {96.00} & {78.41} \\
 & \cellcolor{noise80color}\textbf{\modelname (Ours)} & \cellcolor{noise80color}\textbf{51.28} & \cellcolor{noise80color}\textbf{80.46} &\cellcolor{noise80color} \textbf{88.46} & \cellcolor{noise80color}\textbf{97.49} & \cellcolor{noise80color}\underline{77.04} & \cellcolor{noise80color}\textbf{91.04} & \cellcolor{noise80color}\textbf{96.60} & \cellcolor{noise80color}\underline{78.75} \\
\hline
% --- 80% 噪声块 (浅红色) ---

 & SSN~\cite{ssn}~(AAAI'24) & 20.48 & 43.98 & 54.27 & 74.80 & 56.48 & 77.20 & 89.54 & 50.23 \\

 & CALA~\cite{cala}~(SIGIR'24) & 31.52 & 61.49 & 72.60 & 89.86 & 64.34 & 83.52 & 92.60 & 62.92 \\
 & SPRC~\cite{sprc}~(ICLR'24) & 29.95 & 51.25 & 58.51 & 73.86 & 70.22 & 86.05 & 93.21 & 60.74 \\

 80\% & TME~\cite{TME}~(CVPR'25) & {46.31} & {75.78} & {84.89} & {95.83} & {73.37} & {88.02} & {94.89} & {74.58} \\

& HABIT~\cite{habit} (AAAI'26) & \underline{47.93} & {76.84} & {85.95} & {95.90} & \textbf{74.87} & {89.08} & {95.21} & {75.86} \\

& {INTENT~\cite{intent} (AAAI'26)} & {47.90} & \underline{78.13} & \underline{87.04} & \underline{96.47} & {73.81} & \underline{89.18} & \textbf{95.54} & \underline{75.97} \\
 & \cellcolor{noise80color}\textbf{\modelname (Ours)} & \cellcolor{noise80color}\textbf{48.00} & \cellcolor{noise80color}\textbf{78.53} & \cellcolor{noise80color}\textbf{87.01} & \cellcolor{noise80color}\textbf{97.40} & \cellcolor{noise80color}\underline{74.22} & \cellcolor{noise80color}\textbf{89.52} & \cellcolor{noise80color}\underline{95.42} & \cellcolor{noise80color}\textbf{76.38} \\
\Xhline{1.5pt}
\end{tabular}
}
\vspace{-14pt}  
\label{tab:cirr-noise}
\end{table*}

%%%%%%%%%%%%%%%%%%%%%%%%%%%%%%%%%%%%%%%%%%%%%%%%%%%%%%%%%%%%%%%%%%

\subsection{Performance Comparison}
To systematically evaluate the robustness and generalization capability of \modelname in the NTC scenario, we conduct comprehensive comparative experiments on the FashionIQ and CIRR benchmarks, comparing \modelname with existing traditional CIR baselines and SOTA robust models. As shown in \Cref{tab:fiq_noise} and \Cref{tab:cirr-noise}, our analysis reveals two main findings: 
\textbf{1)} Traditional CIR methods (e.g., SSN~\cite{ssn}, CALA~\cite{cala}, SPRC~\cite{sprc}) are highly sensitive to noise, and their performance declines sharply in high-noise environments. Robust methods significantly outperform traditional methods, and the performance gap widens as noise increases. This highlights the instability of traditional methods in the NTC scenario and emphasizes the necessity of developing robust models.
\textbf{2)} \modelname exhibits the strongest robustness under all noise settings, comprehensively outperforming all SOTA robust baselines, including HABIT~\cite{habit}, INTENT~\cite{intent}, and TME~\cite{TME}. Notably, this performance advantage widens as the noise rate increases: on FashionIQ, \modelname achieves gains over HABIT of $0.92$\% (at 20\% noise), $1.54$\% (at 50\% noise), and $1.41\%$ on the AVG metric. A consistent pattern is observed on the CIRR dataset, with an improvement of +$0.69$\% in the Avg(R@5, R$_{sub}$@1) metric at 80\% noise. These performance gains are attributed to the precise quantification of matching fidelity by Geometric Fidelity Quantization (GFQ), the structured semantic boundaries constructed by Negative Boundary Learning (NBL), and the precise targeted unlearning performed by Boundary-based Targeted Unlearning (BTU). These three components work synergistically to enhance the model's discriminative power and robustness against hard noise during the dynamic optimization process.
%%%%%%%%%%%%%%%%%%%%%%%%%%%%%%%%%%%%%%%%%%%%%%%%%%%%%%%%%%%%%%%%%%

\begin{table}[htbp]
  \centering
  \tabcolsep=9pt
  \caption{Ablation study on FashionIQ and CIRR datasets.}
  \vspace{-8pt}
  \resizebox{\linewidth}{!}{%
    \begin{tabular}{cc|cc|cc}
    \Xhline{1.5pt}
    \multicolumn{1}{c|}{\multirow{2}{*}{D\#}} & \multirow{2}{*}{Deriv.} & \multicolumn{2}{c|}{FashionIQ-Avg} & \multicolumn{2}{c}{CIRR-Avg} \\
\cline{3-6}    \multicolumn{1}{c|}{} &       & R@10  & R@50  & R@K   & R$_{sub}$@K \\
    \hline
    \hline
    \multicolumn{6}{c}{\textit{\textbf{(a) Geometric Fidelity Quantization (GFQ)}}} \\
    \hline
    \multicolumn{1}{c|}{1} & \multicolumn{1}{l|}{w/o random\_target} & 54.25  & 74.45  & 79.01  & 88.23  \\
    \multicolumn{1}{c|}{2} & \multicolumn{1}{l|}{w/o random\_reference} & 54.16  & 74.07  & 80.02  & 88.26  \\
    \multicolumn{1}{c|}{3} & \multicolumn{1}{l|}{w/o random\_mod} & 54.06  & 74.62  & 79.42  & 88.48  \\
    \multicolumn{1}{c|}{4} & \multicolumn{1}{l|}{w/o boundary} & 53.69  & 74.78  & 80.14  & 88.33  \\
    \multicolumn{1}{c|}{5} & \multicolumn{1}{l|}{w/o Fidelity} & 53.42  & 74.09  & 79.90  & 88.20  \\
    \hline
    \multicolumn{6}{c}{\textit{\textbf{(b) Negative Boundary Learning (NBL)}}} \\
    \hline
    \multicolumn{1}{c|}{6} & \multicolumn{1}{l|}{w/o neg-tar\_y} & 53.98  & 74.96  & 79.74  & 88.16  \\
    \multicolumn{1}{c|}{7} & \multicolumn{1}{l|}{w/o neg-tar\_n} & 54.20  & 74.88  & 79.74  & 88.02  \\
    \multicolumn{1}{c|}{8} & \multicolumn{1}{l|}{w/o neg-tar} & 53.62  & 74.40  & 79.74  & 88.06  \\
    \multicolumn{1}{c|}{9} & \multicolumn{1}{l|}{w/o neg-query} & 54.29  & 74.35  & 79.94  & 87.91  \\
    \multicolumn{1}{c|}{10} & \multicolumn{1}{l|}{w/o neg-prompt} & 54.77  & 74.49  & 78.94  & 87.95  \\
    \hline
    \multicolumn{6}{c}{\textit{\textbf{(c) Boundary-based Targeted Unlearning (BTU)}}} \\
    \hline
    \multicolumn{1}{c|}{11} & \multicolumn{1}{l|}{w/o negative} & 54.38  & 74.87  & 79.67  & 87.81  \\
    \multicolumn{1}{c|}{12} & \multicolumn{1}{l|}{w/o OT} & 54.01  & 74.63  & 79.66  & 88.04  \\
    \multicolumn{1}{c|}{13} & \multicolumn{1}{l|}{w/o Unlearn} & 53.13  & 74.02  & 79.72  & 87.75  \\
    \multicolumn{1}{c|}{14} & \multicolumn{1}{l|}{w/o rank} & 52.31  & 73.39  & 79.00  & 87.03  \\
    \hline
    \multicolumn{2}{c|}{\textbf{\modelname (Ours)}} & \textbf{54.93}  & \textbf{75.01}  & \textbf{80.66}  & \textbf{89.06}  \\
    \Xhline{1.5pt}
    \end{tabular}%
    }
    \vspace{-18pt}
  \label{tab:ablation_main}%
\end{table}%

%%%%%%%%%%%%%%%%%%%%%%%%%%%%%%%%%%%%%%%%%%%%%%%%%%%%%%%%%%%%%%%%%%
\subsection{Ablation Studies}

To evaluate the efficacy of each innovative component in \modelname, we conduct a series of comprehensive ablation studies on the FashionIQ and CIRR datasets in \Cref{tab:ablation_main}. We divide these variants into three groups:

\textit{\textbf{G1: Geometric Fidelity Quantization Ablation.}} \textbf{D\#(1): w/o random\_tar}, \textbf{D\#(2): w/o random\_ref}, and \textbf{D\#(3): w/o random\_mod} remove the three types of samples used to estimate the noise boundary $\mathbb{B}$, respectively. \textbf{D\#(4): w/o boundary} removes the boundary $\mathbb{B}$ from the fidelity calculation (Eq.(\ref{eq:fidelity})). \textbf{D\#(5): w/o Fidelity} completely removes the fidelity measure function $\mathcal{F}$ (Eq.(\ref{eq:fidelity})).
\textit{\textbf{G2: Negative Boundary Learning Ablation.}} \textbf{D\#(6): w/o neg-tar\_y} (sets $\textbf{T}$ to all $1$s), \textbf{D\#(7): w/o neg-tar\_n} (sets $\textbf{T}$ to all $-1$s), and \textbf{D\#(8): w/o neg-tar} (sets $\textbf{T}$ to all $0$s) verify the impact of different matching strategies in Target-Oriented Learning. \textbf{D\#(9): w/o neg-query} removes the \textit{Query-oriented Learning} ($\mathcal{L}_{intra}$). \textbf{D\#(10): w/o neg-prompt} removes the core negative prompt vector ($\mathbf{P}_{\text{neg}}$) used for learning the Diagonal Negative Composition ($\mathbf{F}_{neg}$).
\textit{\textbf{G3: Boundary-based Targeted Unlearning Ablation.}} \textbf{D\#(11): w/o negative} disables the effect of the Diagonal Negative Composition ($\mathbf{F}_{neg}$) on Optimal Transport. \textbf{D\#(12): w/o OT} removes the Optimal Transport constraint. \textbf{D\#(13): w/o Unlearn} removes the targeted forgetting loss function ($\mathcal{L}_{ul}$). \textbf{D\#(14): w/o rank} removes the robust contrastive loss ($\mathcal{L}_{robust}$).

We draw the following key conclusions from \Cref{tab:ablation_main}:
\textbf{1)} 
The integrity of the GFQ module is crucial, a fact validated by our ablation study. Systematically removing its core components, whether the random sampling strategy for the noise boundary $\mathbb{B}$ (\textbf{D\#1-D\#3}), the boundary's participation in the fidelity calculation (\textbf{D\#4}), or the entire fidelity measure function $\mathcal{F}$ (\textbf{D\#5}), results in a consistent downward trend across all performance metrics. This strongly indicates each component of GFQ is indispensable for its core design goal of penetrating modality suppression.
\textbf{2)} 
The NBL module's structural integrity is essential, as all ablation variants (\textbf{D\#6-D\#10}) uniformly underperform the full model. The performance drops from removing either \textit{Target-oriented Learning} (\textbf{D\#6-D\#8}) or \textit{Query-oriented Learning} (\textbf{D\#9}) confirms the value of NBL's dual-path constraint for constructing structured negative boundaries.
Notably, the ablation in (\textbf{D\#10}) exhibits the most severe performance drop in this group, particularly on the CIRR dataset. This finding unequivocally demonstrates that explicitly learning the \textit{Diagonal Negative Composition} as a dedicated negative anchor is central to properly anchoring the robust semantic space.
\textbf{3)} 
The mechanisms of the BTU module are highly synergistic. The removal of the foundational Robust Contrastive Loss ($\mathcal{L}_{robust}$) (\textbf{D\#14}) incurs the most significant performance drop among all variants, confirming its fundamental role in suppressing noise and ensuring retrieval accuracy. Concurrently, removing the module's specialized correction components—such as disabling the influence of $\textbf{F}_{neg}$ on OT (\textbf{D\#11}), Optimal Transport itself (\textbf{D\#12}), or the dedicated targeted unlearning loss ($\mathcal{L}_{ul}$) (\textbf{D\#13}), also causes clear performance degradation. This indicates that all parts of the \textit{BTU} mechanism are mutually supportive; the $\mathcal{L}_{robust}$ foundation and the specialized corrective elements (OT, $\mathcal{L}_{ul}$, $\textbf{F}_{neg}$ guidance) are interlocking and collectively form core of targeted unlearning process.

%%%%%%%%%%%%%%%%%%%%%%%%%%%%%%%%%%%%%%%%%%%%%%%%%%%%%%%%%%%%%%%%%%
\subsection{Hyper-parameter sensitivity}

\begin{figure}[ht!]
  \centering
    \vspace{-5pt}
   \resizebox{\linewidth}{!}{\includegraphics[width=\linewidth]{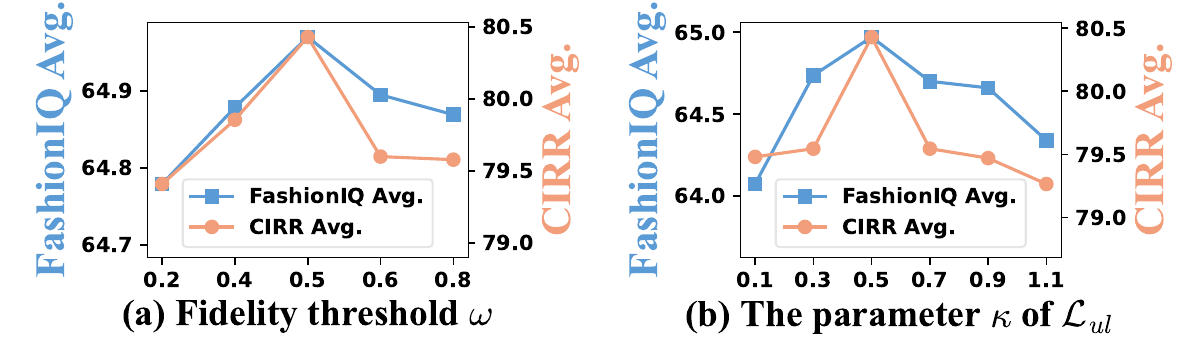}}
      \vspace{-18pt}
   \caption{Sensitivity to (a) Fidelity threshold $\omega$ and (b) $\kappa$ of $\mathcal{L}_{ul}$.}
   \label{fig:sentivity}
   \vspace{-14pt}
\end{figure}

We evaluate the parameter sensitivity of two key hyperparameters $\omega$ and $\kappa$, as shown in \Cref{fig:sentivity}.
\textbf{\textit{a) GFQ fidelity threshold $\omega$.}} As shown in subplot (a), the model performance improves as the value of $\omega$ increases, peaks at $0.5$, and subsequently begins to decline. This trend indicates that excessively low $\omega$ values cause excessive noise samples to be incorrectly classified into the clean set ($\mathcal{T}_{clean}$), which interferes with the core alignment learning. Conversely, excessively high $\omega$ values cause excessive clean samples to be incorrectly classified into the noise set ($\mathcal{T}_{noisy}$), leading them to be excessively unlearned by the BTU module.
\textbf{\textit{b) BTU targeted forgetting loss weight $\kappa$.}} As shown in subplot (b), the model performance steadily increases as the $\kappa$ weight increases, peaks at $0.5$, and subsequently begins to decline. This trend confirms the necessity of the targeted forgetting loss ($\mathcal{L}_{ul}$) weight $\kappa$. A weight that is too low results in insufficient correction for hard noise. Conversely, a weight that is too high produces over-correction when unlearning noisy samples, resulting in Unlearning Backlash, which affects the knowledge of surrounding clean samples.
%%%%%%%%%%%%%%%%%%%%%%%%%%%%%%%%%%%%%%%%%%%%%%%%%%%%%%%%%%%%%%%%%%

\subsection{Case Study} 
\begin{figure}[ht!]
  \centering
    \vspace{-5pt}
   \resizebox{\linewidth}{!}{\includegraphics[width=\linewidth]{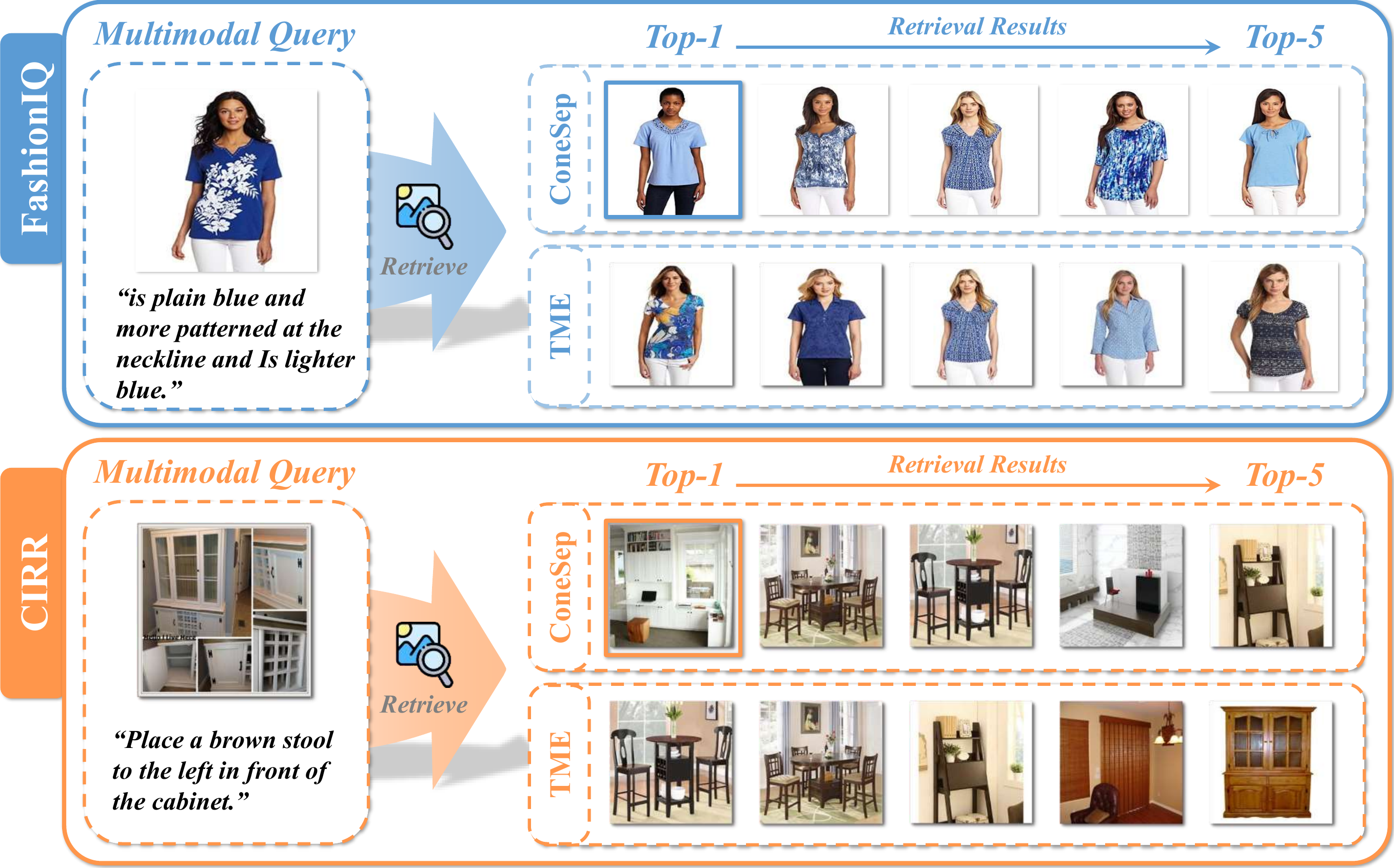}}
      \vspace{-17pt}
   \caption{Case Study on FashionIQ and CIRR.}
   \label{fig:casestudy}
\vspace{-15pt}
\end{figure}

\Cref{fig:casestudy} visually demonstrates ConeSep's retrieval effectiveness, comparing its Top-5 results against the SOTA robust model TME. In the FashionIQ example (\Cref{fig:casestudy}(a)), \modelname successfully retrieves the Top-1 target image matching fine-grained composed semantic requirements (e.g., \textit{a lighter color, a solid body}), whereas TME over-relies on the reference image features. In the CIRR example (\Cref{fig:casestudy}(b)), ConeSep's Top-1 result accurately captures complex spatial relationships, which TME fails to process. This improved performance is attributed to ConeSep's robust framework, which leverages cone effect geometry and targeted unlearning to mitigate noise from the NTC setting. This enhances the model's precision in responding to fine-grained textual constraints without interference from irrelevant reference features.

%%%%%%%%%%%%%%%%%%%%%%%%%%%%%%%%%%%%%%%%%%%%%%%%%%%%%%%%%%%%%%%%%%

%%%%%%%%%%%%%%%%%%%%%%%%%%%%%%%%%%%%%%%%%%%%%%%%%%%%%%%%%%%%%%%%%%

\section{Conclusion}
In this paper, we investigated the NTC problem within the CIR task. We revealed that existing methods, relying on coarse-grained scalar metrics, struggle to address the three challenges of NTC, including \textit{Modality Suppression}, \textit{Negative Anchor Deficiency}, and \textit{Unlearning Backlash}. To address this problem, we proposed ConeSep. It constructed a closed-loop system via its three-stage components, utilizing geometric boundary and diagonal negative composition to achieve precise perception and targeted ``unlearning'' of noise, while leveraging optimal transport to circumvent the unlearning backlash problem. Extensive experiments on two mainstream CIR benchmarks demonstrated that ConeSep significantly outperformed current state-of-the-art techniques across various noise ratios.

\clearpage
\appendix
\setcounter{page}{1}
\setcounter{equation}{0}
\maketitlesupplementary

\noindent This is the appendix of ``ConeSep: Cone-based Robust Noise-Unlearning Compositional Network for Composed Image Retrieval''. 
\begin{itemize}
    \item \textbf{Appendix~\ref{A.3 Sinkhorn-Knopp Algorithm for Targeted Unlearning}}: Sinkhorn-Knopp Algorithm for Targeted Unlearning
    \item \textbf{Appendix~\ref{sup:datasets}}: Datasets   
    \item \textbf{Appendix~\ref{sup:additional_quantitative_analyses}}: Additional Quantitative Analyses
        \begin{itemize}
            \item \textbf{Appendix~\ref{sup:quantitative_efficiency}}: Efficiency Evaluation Analysis
            \item \textbf{Appendix~\ref{sup:quantitative_negative_anchor}}: Quantitative Analysis of Diagonal Negative Composition
            \item \textbf{Appendix~\ref{sup:quantitative_Dhigh_impurity}}: Comparison of Sample Set Partition Purity
            \item \textbf{Appendix~\ref{sup:quantitative_param}}: Additional Hyperparameter Analysis
        \end{itemize}
    \item \textbf{Appendix~\ref{sup:additional_ablation}}: Additional Ablation Study
        \begin{itemize}
            \item \textbf{Appendix~\ref{sup: GFQ Sampling Ablation Studies}} GFQ Sampling Ablation Studies
        \end{itemize}
    \item \textbf{Appendix~\ref{sup:algorithm_training}}: Algorithm of Training Procedure
    \item \textbf{Appendix~\ref{sup:more_qualitative_results}}: More Qualitative Results	
        \begin{itemize}
            \item \textbf{Appendix~\ref{sup:qualitative_negative_anchor_vis}}: Qualitative Analyses of Diagonal Negative Composition
            \item \textbf{Appendix~\ref{sup:qualitative_NTC}}: NTC Identification Analysis
            \item \textbf{Appendix~\ref{sup:qualitative_case}}: More Case Study
        \end{itemize}
\end{itemize}

\section{Sinkhorn-Knopp Algorithm for Targeted Unlearning}
\label{A.3 Sinkhorn-Knopp Algorithm for Targeted Unlearning}
In Section 3.4 of the main text, we formulate the boundary-based targeted unlearning as an entropy-regularized Optimal Transport (OT) problem (Eq. 10). Here, we provide the detailed derivation of the solution using the Sinkhorn-Knopp algorithm.

\textbf{Problem Definition.}
We aim to find the optimal transport plan $\textbf{P}^* \in \mathbb{R}^{B \times 2B}$ that minimizes the transport cost regarding $\textbf{C}_{masked}$ while maximizing entropy. Following the notation in the main text, the objective is:
\begin{equation}
    \textbf{P}^* = \arg\min_{\textbf{P} \in \Pi(\textbf{u}, \textbf{v})} s(\textbf{P}, \textbf{C}_{masked}) - \epsilon H(\textbf{P}),
\end{equation}
where $\small s(\textbf{P}, \textbf{C}_{masked}) = \langle \textbf{P}, \textbf{C}_{masked} \rangle = \sum_{ij} \textbf{P}_{ij} [\textbf{C}_{masked}]_{ij}$ represents the total transport cost.
The feasible set $\Pi(\textbf{u}, \textbf{v})$ is defined by the marginal constraints:
\begin{equation}
    \Pi(\textbf{u}, \textbf{v}) = \{ \textbf{P} \in \mathbb{R}_+^{B \times 2B} \mid \textbf{P}\textbf{1}_{2B} = \textbf{u}, \textbf{P}^\top\textbf{1}_{B} = \textbf{v} \},
\end{equation}
where $\textbf{u} \in \mathbb{R}^B$ and $\textbf{v} \in \mathbb{R}^{2B}$ are the source and target marginal distributions (typically uniform, i.e., $\textbf{u} = \frac{1}{B}\textbf{1}_B, \textbf{v} = \frac{1}{2B}\textbf{1}_{2B}$).

\textbf{Lagrangian and Gibbs Kernel.}
To solve this constrained optimization problem, we introduce Lagrange multipliers $\boldsymbol{\alpha} \in \mathbb{R}^B$ and $\boldsymbol{\beta} \in \mathbb{R}^{2B}$ for the marginal constraints. The Lagrangian is given by:
\begin{equation}
\begin{split}
    \mathcal{L}(\textbf{P}, \boldsymbol{\alpha}, \boldsymbol{\beta}) &= \sum_{ij} \textbf{P}_{ij} \textbf{C}_{ij} + \epsilon \sum_{ij} \textbf{P}_{ij}(\log \textbf{P}_{ij} - 1) 
    \\&+ \boldsymbol{\alpha}^\top (\textbf{P}\textbf{1} - \textbf{u}) + \boldsymbol{\beta}^\top (\textbf{P}^\top\textbf{1} - \textbf{v}).
\end{split}
\end{equation}
Taking the derivative with respect to $\textbf{P}_{ij}$ and setting it to zero:
\begin{equation}
    \frac{\partial \mathcal{L}}{\partial \textbf{P}_{ij}} = \textbf{C}_{ij} + \epsilon \log \textbf{P}_{ij} + \boldsymbol{\alpha}_i + \boldsymbol{\beta}_j = 0.
\end{equation}
Solving for $\textbf{P}_{ij}$, we obtain the form of the optimal solution:
\begin{equation}
    \textbf{P}_{ij}^* = \exp\left( -\frac{\boldsymbol{\alpha}_i}{\epsilon} \right) \exp\left( -\frac{\textbf{C}_{ij}}{\epsilon} \right) \exp\left( -\frac{\boldsymbol{\beta}_j}{\epsilon} \right).
\end{equation}
We define the \textbf{Gibbs Kernel matrix} $\textbf{K}$ as $\textbf{K}_{ij} = \exp( -[\textbf{C}_{masked}]_{ij} / \epsilon )$.
By letting the scaling vectors be $\textbf{a}_i = \exp(-\boldsymbol{\alpha}_i/\epsilon)$ and $\textbf{b}_j = \exp(-\boldsymbol{\beta}_j/\epsilon)$, the optimal plan factorizes into:
\begin{equation}
    \textbf{P}^* = \text{diag}(\textbf{a}) \textbf{K} \text{diag}(\textbf{b}).
\end{equation}

\textbf{Iterative Scaling Updates.}
The unknown scaling vectors $\textbf{a}$ and $\textbf{b}$ must be determined such that $\textbf{P}^*$ satisfies the marginal constraints. Substituting the factorization into the constraints yields:
\begin{equation}
    \textbf{a} \odot (\textbf{K}\textbf{b}) = \textbf{u}, \quad \textbf{b} \odot (\textbf{K}^\top\textbf{a}) = \textbf{v}.
\end{equation}
These equations are solved via the Sinkhorn-Knopp fixed-point iteration. In each step $t$, we update:
\begin{align}
    \textbf{a}^{(t+1)} &\leftarrow \frac{\textbf{u}}{\textbf{K} \textbf{b}^{(t)}}, \\
    \textbf{b}^{(t+1)} &\leftarrow \frac{\textbf{v}}{\textbf{K}^\top \textbf{a}^{(t+1)}},
\end{align}
where division is element-wise. The algorithm typically converges within 10-20 iterations. The final smooth transport plan $\textbf{P}^*$ is then used as the soft target $Y$ (Eq. 11) to guide the targeted unlearning.

%%%%%%%%%%%%%%%%%%%%%%%%%%%%%%%%%%%%%%%%%%%%%
\section{Datasets}
\label{sup:datasets}

We evaluate the performance of \modelname on two widely recognized benchmarks. The detailed descriptions are as follows:

\begin{itemize}
    \item \textbf{FashionIQ}~\cite{FashionIQ} is a premier CIR benchmark dedicated to the fashion domain. It comprises 77,684 high-resolution fashion images and 30,134 annotated triplets. These triplets are explicitly categorized into three distinct subsets: \textit{Dresses}, \textit{Shirts}, and \textit{Tops\&Tees}. The primary challenge of this dataset lies in its modification texts, which focus on describing fine-grained visual attribute manipulations (e.g., ``change to V-neck'' or ``denser stripes''), requiring the model to capture subtle semantic differences.
    \item \textbf{CIRR}~\cite{cirr} is a large-scale, open-domain benchmark with images derived from the classic NLVR2~\cite{NLVR2} dataset. It contains a total of 21,552 unique images and 36,554 annotated triplets. In contrast to FashionIQ, CIRR presents distinct challenges: (1) its images depict complex real-world scenes containing multiple objects; (2) the modification texts involve not only attribute changes but also diverse compositional logic such as spatial relationships and object addition/deletion. Furthermore, to mitigate the issue of false negatives, CIRR provides a specific candidate subset for each query during evaluation.
\end{itemize}

\section{Additional Quantitative Analysis}
\label{sup:additional_quantitative_analyses}

\subsection{Efficiency Evaluation Analysis}
\label{sup:quantitative_efficiency}

To comprehensively evaluate the feasibility and cost-effectiveness of the model in practical scenarios, we employed multi-dimensional evaluation metrics in~\Cref{tab:efficiency} to conduct a detailed comparison of various methods. Specifically, \textbf{FLOPs(G)} quantifies the theoretical computational complexity during a single forward pass, while \textbf{Parameters(M)} reflects the model's parameter size and storage footprint. Regarding time and resource dimensions, \textbf{Test time(s/sample)} records the average latency for processing a single query sample during inference, directly correlating with the real-world deployment experience; \textbf{Train Time(s/iteration)} measures the time cost per iteration during training; and \textbf{GPU Memory(MiB)} indicates the peak GPU memory usage under a fixed batch size, reflecting hardware entry barriers. Furthermore, we incorporate \textbf{FashionIQ-Avg} and \textbf{CIRR-Avg} as core performance indicators to comprehensively assess the Efficiency-Performance Trade-off.

Comparative analysis based on the aforementioned metrics demonstrates that \modelname achieves improvements in both inference efficiency and retrieval performance while maintaining a comparable parameter scale and computational load. Regarding model complexity, \modelname's parameter count ($915.69$M) and FLOPs ($411.51$G) are comparable to the ordinary baseline SPRC and the robust baseline TME, indicating that the performance gains do not stem from a blind expansion of model scale. Notably, \modelname exhibits superior efficiency during the inference phase, with a single-sample test time of only $0.0091$ seconds. Benefiting from our core modules (GFQ, NBL, BTU) applying constraints solely during training without participating in inference calculations, \modelname's inference speed is slightly faster than SPRC ($0.011$s) and approximately $13.6\times$ faster than TME ($0.124$s). Regarding training overhead, although introducing robustness constraints results in higher GPU memory usage ($25807$ MiB) and iteration time ($5.15$s) for \modelname compared to the simple SPRC, our training efficiency is improved by approximately $34.5\%$ compared to the robust counterpart TME. Crucially, \modelname exchanges reasonable training resources for significant performance returns, achieving the best retrieval accuracy on both benchmark datasets (FashionIQ-Avg=$64.97$\%, CIRR-Avg=$80.43$\%). In summary, \modelname successfully unifies SOTA-level robustness with inference efficiency, demonstrating substantial value for real-world deployment.
%%%%%%%%%%%%%%%%%%%%%%%%%%%%%%%%%%%%%%%%%%%%%

\begin{table*}[ht]
  \centering
  \caption{Comparison of computational complexity and efficiency among SPRC, TME, and ConeSep with its ablation variants.}
    \resizebox{\linewidth}{!}{%
    \begin{tabular}{c|c|c|c|c|c|c|c|c}
    \Xhline{1.5pt}
    Type  & Method & FLOPs(G) & Parameters(M) & GPU Memory(MiB) & Test time(s/sample) & Train Time(s/iteration) & FashionIQ-Avg & CIRR-Avg \\
    \hline
    Ordinary & SPRC  & 413.38 & 915.69 & 24478(bs=128) & 0.011 & \textbf{2.624(bs=128)} & 56.33 & 76.98 \\
    \hline
    \multirow{2}{*}{Robust} & TME   & \textbf{405.2} & \textbf{915.68} & \textbf{12405(bs=128)} & 0.124 & 7.858(bs=128) & 63.97 & 79.74 \\
          & \textbf{\modelname (Ours)} & 411.51     & 915.69     & 25807(bs=128)     & \textbf{0.0091}     & 5.15(bs=128)     & \textbf{64.97}     & \textbf{80.43} \\
    \Xhline{1.5pt}
    \end{tabular}%
    }
  \label{tab:efficiency}%
\end{table*}%

\subsection{Quantitative Analysis of Diagonal Negative Composition}
% \subsection{负向锚点定量}
\label{sup:quantitative_negative_anchor}

\begin{figure}
    \centering
    \includegraphics[width=\linewidth]{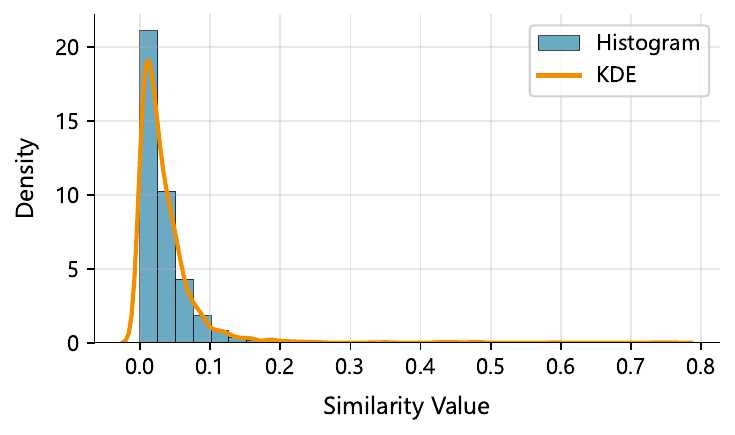}
    \caption{Visualization of the cosine similarity distribution between the composed feature $\mathbf{F}_c$ and the learned diagonal negative composition $\mathbf{F}_{neg}$. The distribution is highly concentrated around 0, indicating an orthogonal relationship.}
    \label{fig:neg_anchor_dist}
\end{figure}

In order to further verify whether the ``Diagonal Negative Composition'' ($\mathbf{F}_{neg}$) constructed by the Negative Boundary Learning (NBL) module truly serves as an effective negative anchor, we perform quantitative analyses on its geometric separability from the original composed feature ($\mathbf{F}_c$). Specifically, we calculate the cosine similarity distribution for all sample pairs $(\mathbf{F}_c, \mathbf{F}_{neg})$ on the FashionIQ validation set and visualize it as a histogram and a Kernel Density Estimation (KDE) curve, as shown in~\Cref{fig:neg_anchor_dist}. Observing this figure, we find that the similarity values exhibit extremely high concentration. The similarities for the vast majority of samples are tightly distributed within the narrow range of $[0, 0.1]$, and the distribution peak significantly approaches $0$. This statistical result strongly demonstrates that $\mathbf{F}_{neg}$ maintains a strictly Approximate Orthogonality relationship with $\mathbf{F}_c$ in the high dimensional feature space. This phenomenon is consistent with the constraint optimization objective that we design through $\mathcal{L}_{intra}$ in the methodology section (Section~\ref{subsubsec:Negative Boundary Learning}), which is to force the learned negative anchor to be orthogonal to the query in terms of semantic direction. This orthogonality ensures that $\mathbf{F}_{neg}$ neither contains the query's effective semantics (thus avoiding the loss of positive knowledge) nor overlaps with the target image features, thereby successfully constructing a clear, independent structured boundary that provides precise directional guidance for the Targeted Unlearning (BTU) module.

\subsection{Comparison of Sample Set Partition Purity}
% \subsection{样本集合划分纯度对比}
\label{sup:quantitative_Dhigh_impurity}

In Section \ref{subsubsec:Geometric Fidelity Quantization}, we propose the Geometric Fidelity Quantization (GFQ) module, which aims to distinguish the high-fidelity clean set ($\mathcal{T}_{clean}$) from the low-fidelity noisy set ($\mathcal{T}_{noisy}$) by estimating geometric noise boundaries. However, our subsequent Boundary-based Targeted Unlearning (BTU) module does not directly treat the filtered $\mathcal{T}_{clean}$ as absolute Ground Truth. Instead, it employs an optimal transport (OT) based soft correction mechanism. The core premise of this design is that, constrained by the ``Modality Suppression'' challenge, a portion of ``hard noise'' samples are theoretically inevitable within $\mathcal{T}_{clean}$. These hard noise samples are instances where the reference image and the target image exhibit high visual similarity, but the modification text is incorrect. To validate this premise and simultaneously evaluate the efficacy of our filtering strategy compared to existing methods, we conduct an in-depth quantitative analyses of the purity of $\mathcal{T}_{clean}$ in this section.

\begin{figure}[h]
  \centering
    \vspace{-5pt}
   \resizebox{\linewidth}{!}{\includegraphics[width=\linewidth]{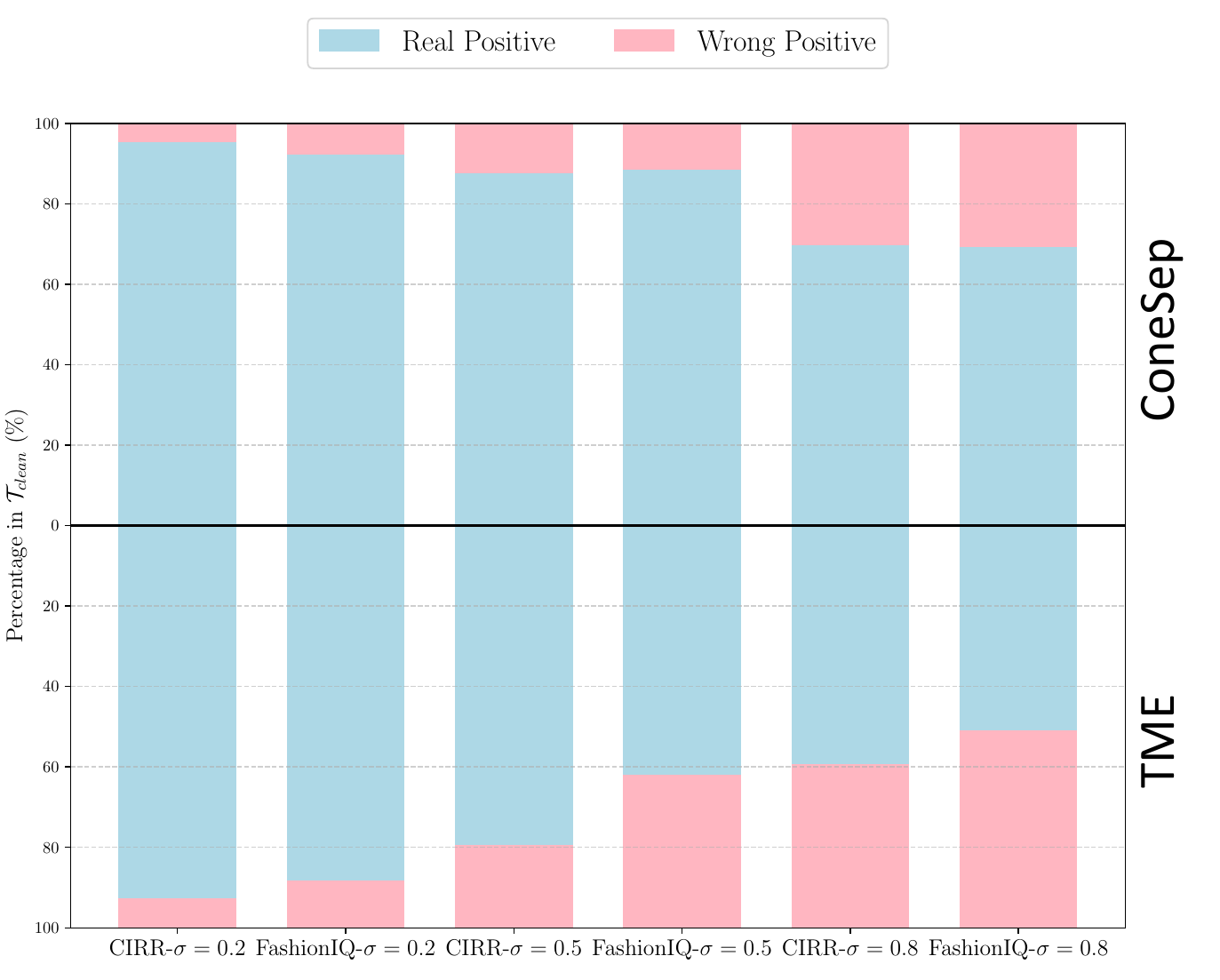}}
     \vspace{-18pt}
   \caption{Quantitative analyses and comparison of the internal purity of the $\mathcal{T}_{clean}$ set.
  This figure demonstrates the composition of the "high-fidelity" set selected by the model under different initial noise rates $\sigma$.
  The blue bars (Real Positive) represent the genuine clean samples correctly identified, while the pink bars (Wrong Positive) represent the hard noise mistakenly classified as clean samples.
  The upper part shows our \textbf{ConeSep} model, and the lower part shows the \textbf{TME} baseline model.
  A smaller proportion of the pink area indicates a higher purity of the selected set. The results show that \modelname maintains a high purity even under high noise and significantly outperforms TME.}
  \label{fig:impurity_stack}
  \vspace{-14pt}
\end{figure}

\begin{figure*}[ht!]
  \centering
    \vspace{-5pt}
   \resizebox{\linewidth}{!}{\includegraphics[width=\linewidth]{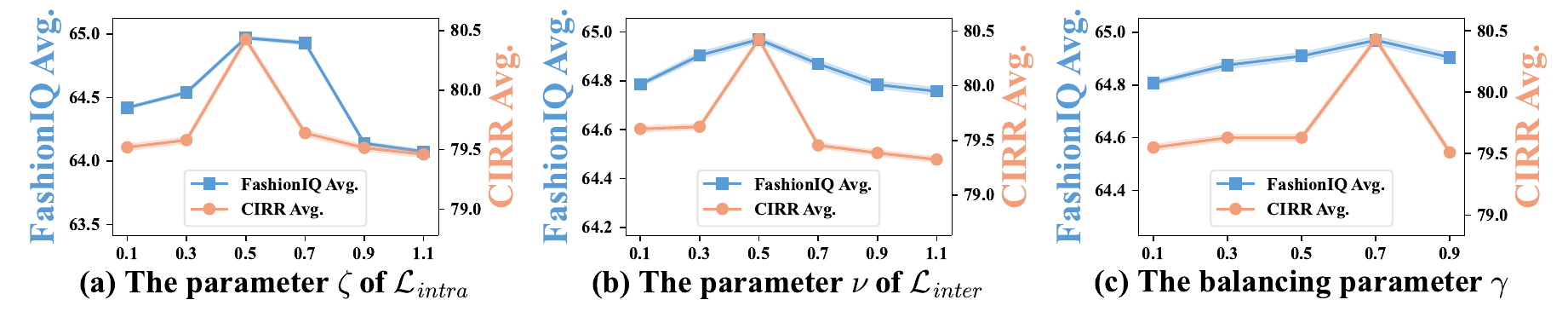}}
      \vspace{-18pt}
   \caption{Sensitivity analysis of hyperparameters on FashionIQ and CIRR datasets: (a) the intra-modal loss weight $\zeta$, (b) the inter-modal loss weight $\nu$, and (c) the balancing parameter $\gamma$.}
   \label{fig:more_sentivity}
   \vspace{-4pt}
\end{figure*}

Since real-world datasets lack Ground Truth annotations for Noisy Triplet Correspondence (NTC), we conduct experiments on the FashionIQ and CIRR datasets using synthetic noise settings based on $\sigma \in \{0.2, 0.5, 0.8\}$. This setup allows us to precisely track which samples are artificially injected noise. Figure \ref{fig:impurity_stack} intuitively illustrates the composition of the sample set identified by the model as high fidelity. The blue region represents the genuinely clean samples that are correctly identified (Real Positive), while the pink region represents the noise triplets that are incorrectly classified as clean samples (Wrong Positive, i.e., hard noise). The upper part of the figure shows the performance of our \modelname at different noise rates, and the lower part shows the performance of TME~\cite{TME}, the current state-of-the-art robust baseline model.

By observing Figure \ref{fig:impurity_stack}, we clearly find a significant difference in noise resistance between ConeSep (upper part) and the baseline method TME (lower part). TME exhibits noticeable performance degradation in high-noise environments. In the $\sigma=0.8$ setting, the pink region in its filtered set expands, even approaching the proportion of the blue region on FashionIQ. This indicates that the Gaussian Mixture Model (GMM) upon which TME relies cannot effectively distinguish samples that are visually similar but semantically inconsistent when facing substantial hard noise. In contrast, our ConeSep demonstrates superior robustness. Benefiting from the effective penetration of geometric noise boundary against Modality Suppression, the blue region remains dominant in ConeSep's filtering results, even under extremely high noise interference. This strongly proves the effectiveness of the GFQ module in decouple features and noise identification, and the purity of its filtered samples is significantly better than existing SOTA methods.

Although ConeSep's filtering purity is clearly superior to others, a small amount of pink residue is still observable in the figure. This objectively reflects the inherent challenge of ``hard noise'' in the NTC task: some samples are difficult to be completely ``hard'' cut off by any filter due to their extremely high visual similarity. This observation provides solid empirical evidence for the Boundary-based Targeted Unlearning (BTU) module designed in Section \ref{subsubsec:Boundary-based Targeted Unlearning}. Precisely because we recognize that Hard Filtering cannot theoretically achieve 100\% perfection, we do not directly treat $\mathcal{T}_{clean}$ as absolute Ground Truth. Instead, we introduce an optimal transport-based soft correction mechanism. This mechanism robustly tolerates and handles the residual impurities that are unavoidable during the filtering stage, thereby maximizing the reduction of the risk of overfitting the residual noise while utilizing the clean sample information, achieving training robustness in a noisy environment.

\subsection{Additional Hyperparameter Analysis}
\label{sup:quantitative_param}

%%%%%%%%%%%%%%%%%%%%%%%%%%%%%%%%%%%%%%%%%%%%%
%
In this section, we further discuss the parameter sensitivity of three additional crucial hyperparameters in ConeSep: the intra-modal loss weight $\zeta$ and inter-modal loss weight $\nu$ within the NBL module, and the balancing parameter $\gamma$ within the BTU module. We evaluate the impact of these parameters on the FashionIQ and CIRR datasets, and the results are illustrated in~\Cref{fig:more_sentivity}.

\textbf{\textit{a) Intra-modal loss weight $\zeta$.}} As shown in subplot (a), the model performance exhibits a trend of increasing initially and then decreasing as the value of $\zeta$ increases, peaking at $0.5$. $\zeta$ controls the weight of $\mathcal{L}_{intra}$, a loss designed to constrain the similarity between the composed feature $\mathbf{F}_c$ and its diagonal negative composition $\mathbf{F}_{neg}$, ensuring they remain orthogonal in the metric space. When $\zeta$ is too low, the model fails to effectively shape $\mathbf{F}_{neg}$ as the opposite of the query semantics, resulting in a blurred negative boundary. Conversely, when $\zeta$ is too high, the enforced orthogonal constraint may distort the geometric structure of the feature space, thereby interfering with the core retrieval paradigm learning in the Positive Alignment Path.

\textbf{\textit{b) Inter-modal loss weight $\nu$.}} As shown in subplot (b), the model achieves optimal performance at $\nu=0.5$. $\nu$ regulates the weight of $\mathcal{L}_{inter}$, which is responsible for pushing $\mathbf{F}_{neg}$ away from its corresponding target image $\mathbf{F}_t$ through \textit{Target-oriented Learning}. This trend indicates that a moderate $\nu$ value facilitates the construction of a high-quality Diagonal Negative Composition, serving as an effective directional guide. However, an excessively high $\nu$ causes the optimization process to over-focus on constructing the negative boundary, distracting the model from pulling positive pairs closer via $\mathcal{L}_{rank}$, thus compromising the final retrieval accuracy.

\textbf{\textit{c) The balancing parameter $\gamma$.}} As shown in subplot (c), the model performance improves steadily as $\gamma$ increases from $0.1$ to $0.7$, followed by a decline. $\gamma$ balances the global optimal plan $\mathbf{P}$ derived from Optimal Transport (OT) and the original hard label $\mathbf{L}$ when constructing the smooth soft label $\mathbf{Y}$. A lower $\gamma$ implies that the model over-relies on the hard label $\mathbf{L}$, ignoring the geometric structural information provided by OT, which leads to a lack of smoothness in the unlearning process. On the other hand, an overly high $\gamma$ may weaken the explicit ``targeted unlearning'' instruction for noisy samples inherent in the hard labels (i.e., the strong supervisory signal in $\mathbf{L}$ where the diagonal for noisy samples is set to $0$), resulting in insufficient correction strength for the noise.
%%%%%%%%%%%%%%%%%%%%%%%%%%%%%%%%%%%%%%%%%%%%%

\section{Additional Ablation Study}
\label{sup:additional_ablation}

\begin{table*}[ht]
  \centering
  \tabcolsep=10pt
  \caption{Ablation study on different sampling strategies for boundary estimation on FashionIQ ($\sigma$=0.2).}
  \vspace{-8pt}
  \resizebox{\linewidth}{!}{%
    \begin{tabular}{c|cc|cc|cc|cc|c}
    \Xhline{1.5pt}
    \multirow{2}{*}{Strategy} & \multicolumn{2}{c|}{Dress} & \multicolumn{2}{c|}{Shirt} & \multicolumn{2}{c|}{Toptee} & \multicolumn{3}{c}{Average} \\
\cline{2-10}    \multicolumn{1}{c|}{} & R@10  & R@50  & R@10  & R@50  & R@10  & R@50  & R@10  & R@50  & AVG. \\
    \hline
    \hline
    \multicolumn{1}{c|}{w/ Empirical} & 49.55  & 72.10  & 55.02  & 74.35  & 57.50  & 77.05  & 54.02  & 74.50  & 64.26  \\
    \multicolumn{1}{c|}{w/ Uniform Dist} & 49.21  & 71.85  & 54.86  & 74.01  & 57.33  & 76.82  & 53.80  & 74.22  & 64.01  \\
    \multicolumn{1}{c|}{w/ Laplace Dist} & 49.80  & 72.45  & 55.15  & 74.20  & 57.65  & 77.20  & 54.20  & 74.62  & 64.41  \\
    \hline
    \multicolumn{1}{c|}{\textbf{\modelname (Gaussian)}} & \textbf{50.23}  & \textbf{72.19}  & \textbf{56.28}  & \textbf{74.45}  & \textbf{58.29}  & \textbf{78.38}  & \textbf{54.93}  & \textbf{75.01}  & \textbf{64.97}  \\
    \Xhline{1.5pt}
    \end{tabular}%
    }
  \label{tab:sampling_strategy_fiq}%
\end{table*}%

\begin{table*}[ht]
  \centering
    \tabcolsep=12pt
  \caption{Ablation study on different sampling strategies for boundary estimation on CIRR ($\sigma$=0.2).}
  \vspace{-8pt}
  \resizebox{\linewidth}{!}{%
    \begin{tabular}{c|cccc|ccc|c}
    \Xhline{1.5pt}
    \multirow{2}{*}{Strategy} & \multicolumn{4}{c|}{R@K} & \multicolumn{3}{c|}{R$_{sub}$@K} & \multirow{2}{*}{Avg} \\
\cline{2-9}    \multicolumn{1}{c|}{} & K=1 & K=5 & K=10 & K=50 & K=1 & K=2 & K=3 & \\
    \hline
    \hline
    \multicolumn{1}{c|}{w/ Empirical} & 51.65 & 81.50 & 89.45 & 97.80 & 77.55 & 91.20 & 96.45 & 79.85 \\
    \multicolumn{1}{c|}{w/ Uniform Dist} & 51.30 & 81.25 & 89.10 & 97.65 & 77.20 & 90.95 & 96.30 & 79.65 \\
    \multicolumn{1}{c|}{w/ Laplace Dist} & 51.95 & 81.85 & 89.70 & 98.05 & 77.90 & 91.45 & 96.60 & 80.15 \\
    \hline
    \multicolumn{1}{c|}{\textbf{\modelname (Gaussian)}} & \textbf{52.29} & \textbf{82.19} & \textbf{89.98} & \textbf{98.19} & \textbf{78.66} & \textbf{91.76} & \textbf{96.75} & \textbf{80.43} \\
    \Xhline{1.5pt}
    \end{tabular}%
    }
  \label{tab:sampling_strategy_cirr}%
\end{table*}%

\subsection{GFQ Sampling Ablation Studies}
\label{sup: GFQ Sampling Ablation Studies}

\statement{Effectiveness of Gaussian Sampling Strategy.} 
Impact of Boundary Sampling Strategies. To strictly validate the theoretical premise of our \textit{Geometric Fidelity Quantization}, we conduct a comprehensive comparison of boundary estimation strategies on both FashionIQ and CIRR datasets. As shown in Table~\ref{tab:sampling_strategy_fiq} and Table~\ref{tab:sampling_strategy_cirr}, we compare our Gaussian strategy with three alternatives: (1) \textbf{w/ Empirical}, which derives the boundary from the shuffled training batch; (2) \textbf{w/ Uniform Dist}, which samples vectors from a bounded Uniform distribution; and (3) \textbf{w/ Laplace Dist}, which samples from a Laplace distribution characterized by sharper peaks and heavier tails.

The results across both benchmarks consistently demonstrate that \modelname (Gaussian) yields the best performance (e.g., 65.31\% Avg on FashionIQ and 80.43\% Avg on CIRR). Specifically, w/ Empirical exhibits performance degradation due to the high variance of mini-batch statistics, which leads to unstable boundary estimation. Furthermore, while w/ Laplace Dist. generally outperforms w/ Uniform Dist. by better approximating the central tendency of high-dimensional features, it still falls short of the Gaussian strategy. This empirical evidence confirms that the aggregated deep embeddings asymptotically converge to a Gaussian distribution. Consequently, our Gaussian sampling strategy provides the most accurate and stable geometric estimation for the noise boundary in the metric space.

\begin{figure}[ht]
  \centering
    \vspace{-5pt}
   \resizebox{\linewidth}{!}{\includegraphics[width=0.8\linewidth]{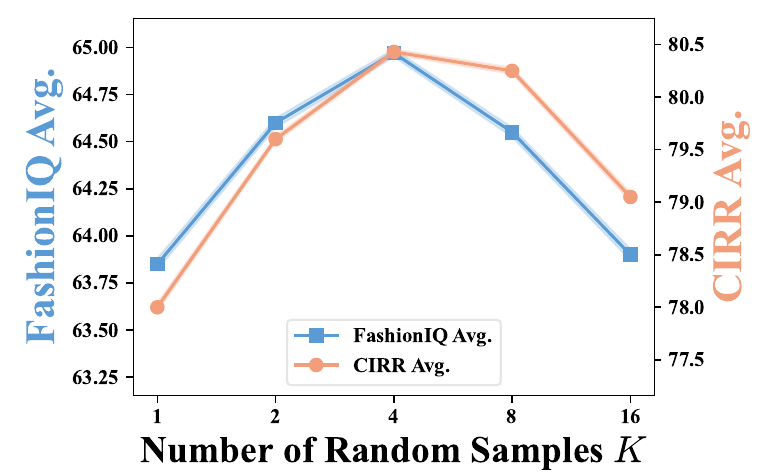}}
      \vspace{-18pt}
   \caption{Sensitivity analysis of the number of random samples on FashionIQ and CIRR datasets.}
   \label{fig:sensitivity_k}
   \vspace{-4pt}
\end{figure}

\statement{Number of Random Samples $K$ of GFQ.} 
Figure~\ref{fig:sensitivity_k} illustrates the impact of the number of random samples $K$ on the model's performance. 
As $K$ increases from 1 to 4, we observe a consistent improvement in retrieval accuracy on both FashionIQ and CIRR datasets. 
This trend validates that a single sample ($K=1$) is insufficient to robustly estimate the geometric noise boundary $\mathbb{B}$, as it is susceptible to random fluctuations in the high-dimensional space. 
Increasing $K$ allows for a more accurate approximation of the boundary's expectation, thereby stabilizing the Fidelity Quantization process.
However, the performance saturates and exhibits a slight decline when $K$ exceeds 4. 
This suggests that $K=4$ provides a sufficient statistical sample to capture the distribution characteristics of the noise boundary. 
Further increasing $K$ yields diminishing returns and may lead to an overly smoothed boundary estimation that lacks the flexibility to handle specific hard samples, while also incurring unnecessary computational overhead. 
Therefore, we adopt $K=4$ as the optimal setting to balance estimation stability and computational efficiency.

\section{Algorithm of Training Procedure}
\label{sup:algorithm_training}

To support the methodology discussion in the main text and comprehensively display the implementation logic of \modelname, we provide detailed training algorithm pseudocode in \textbf{Algorithm~\ref{alg:conesept_training}}. This training process constructs a closed-loop optimization system that organically integrates Geometric Fidelity Quantization (GFQ), Negative Boundary Learning (NBL), and Boundary-based Targeted Unlearning (BTU) into a unified training pipeline.

The execution of the algorithm initiates with a feature extraction and perception phase, where the model utilizes Gaussian random sampling to dynamically estimate the geometric noise boundary $\mathbb{B}$. This process quantifies the matching fidelity of samples and explicitly partitions the current batch into a high-fidelity clean set $\mathcal{T}_{\text{clean}}$ and a low-fidelity noisy set $\mathcal{T}_{\text{noisy}}$. Subsequently, the training adheres to a progressive dual-stage strategy. During the initial warm-up phase (the first $N$ Epochs), the algorithm focuses on the construction of the NBL path. By optimizing the Target-oriented and Query-oriented loss functions, the model constructs a robust ``Diagonal Negative Composition'' $\mathbf{F}_{\text{neg}}$ for each query to serve as a semantic anchor. Upon the establishment of the boundaries, the process transitions smoothly to the joint optimization phase based on BTU. In this phase, the algorithm employs a specifically designed mask matrix $\mathbf{M}$, which blocks the positive path of noisy samples while preserving the negative path of clean samples, to solve the entropy-regularized optimal transport problem. This operation generates smooth soft labels $\mathbf{Y}$ to execute precise Targeted Unlearning. Ultimately, the entire framework undergoes end-to-end updates by jointly minimizing the Robust Contrastive Loss, Query-oriented Learning loss, and targeted forgetting loss. Consequently, the model achieves robust and accurate Composed Image Retrieval (CIR) within complex Noisy Triplet Correspondence (NTC) environments.

\section{More Qualitative Results}
\label{sup:more_qualitative_results}

\subsection{Qualitative Analyses of Diagonal Negative Composition}
\label{sup:qualitative_negative_anchor_vis}

To intuitively verify whether our proposed ``Diagonal Negative Composition'' ($\mathbf{F}_{neg}$) truly learns the semantic opposition required as a ``negative anchor'', we conduct visualized comparative retrieval experiments. Specifically, for a given query $\langle x_r, x_m \rangle$, we use the positive composed feature $\mathbf{F}_c$ generated by the model and the Diagonal Negative Composition $\mathbf{F}_{neg}$ to perform nearest neighbor searches in the retrieval gallery and display the Top-5 retrieval results. In this context, the ``semantic opposition'' we define does not refer to completely unrelated random images but specifically refers to a semantic state that violates or ignores the instructions of the modification text $x_m$. For example, if the $x_m$ instruction is ``change to red'', then ``keeping the original blue'' or ``wrong green'' constitutes ``hard negative'' semantics that are not only visually confusing but also logically contrary to the instruction.

\begin{figure*}[ht!]
  \centering
    \vspace{-5pt}
   \resizebox{\linewidth}{!}{\includegraphics[width=\linewidth]{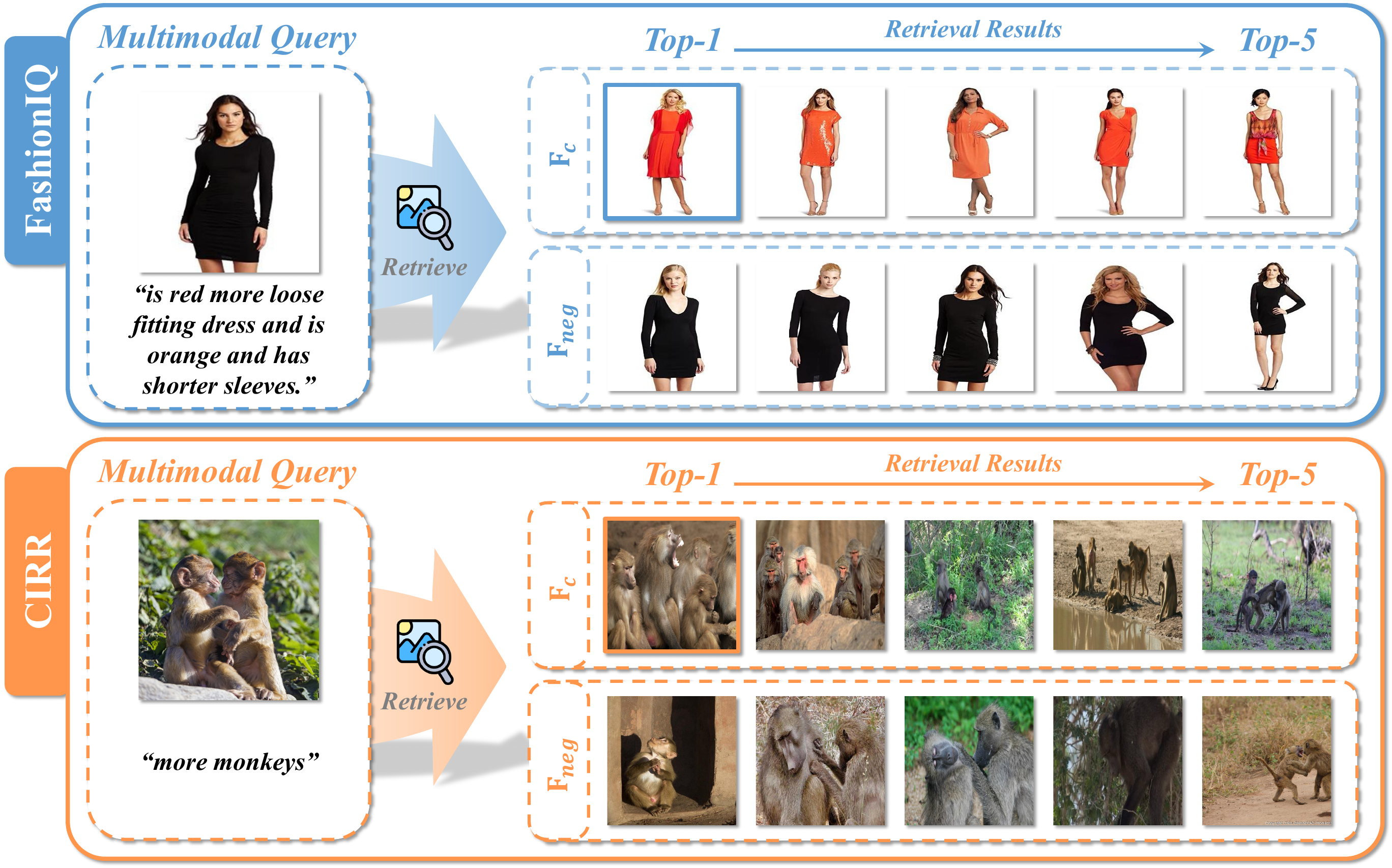}}
      \vspace{-18pt}
   \caption{
  Visualization of retrieval results using the composed feature $\mathbf{F}_c$ and the diagonal negative composition $\mathbf{F}_{neg}$.
  \textbf{Top (FashionIQ):} While $\mathbf{F}_c$ correctly retrieves red dresses following the text instruction, $\mathbf{F}_{neg}$ retrieves black dresses that retain the reference visual semantics but violate the modification logic.
  \textbf{Bottom (CIRR):} $\mathbf{F}_c$ retrieves images with ``more monkeys'', whereas $\mathbf{F}_{neg}$ retrieves images with fewer or single monkeys, effectively capturing the semantic opposite of the instruction.
  }
   \label{fig:supp_neg_vis}
   \vspace{-14pt}
\end{figure*}

As shown in~\Cref{fig:supp_neg_vis}, the visualization results clearly reveal the significant differences in semantic direction between $\mathbf{F}_{neg}$ and $\mathbf{F}_c$. In the case of FashionIQ (top row), the reference image is a black tight dress, and the modification text requires changing it to ``red, looser, and with shorter sleeves''. Observing the retrieval results reveals that the results of $\mathbf{F}_c$ accurately align with the target semantics, and the Top-5 images are all red dresses conforming to the description. In sharp contrast, the results retrieved using $\mathbf{F}_{neg}$ mainly consist of black dresses with similar styles. This indicates that $\mathbf{F}_{neg}$ successfully captures the semantic state of ``ignoring text instructions'', which implies retaining the visual features of the reference image such as the black color while failing to execute the color conversion. This state represents the most typical ``hard noise'' in the Noisy Triplet Correspondence (NTC) problem, characterized by visual similarity but semantic mismatch, and serves as the direction that the model aims to ``push away'' in the Boundary-based Targeted Unlearning (BTU) module.

\begin{algorithm}[H] % 注意：去掉了 *，改为单栏环境
\caption{Training Procedure of ConeSep}
\label{alg:conesept_training}
\textbf{Input}: Reference image $x_r$, target image $t$, modification text $y_m$. \\
\textbf{Parameters}: Total epochs $N_{{total}}$, NBL epochs $N$, Learning Rate $\eta$, Fidelity threshold $\omega$, Loss weights ($\zeta, \kappa, \nu, \gamma$). \\
\textbf{Output}: Fine-tuned model parameters $\Psi^*$. \\
\begin{algorithmic}[1]
\REQUIRE Training Dataset $\mathcal{D}$, Batch size $B$.
\STATE Initialize all model components $\Psi$.
\FOR{$epoch = 1$ to $N_{total}$}
\FOR{batch $\mathcal{B}$ in $\mathcal{D}$}
\STATE // \textbf{1. Feature Extraction and Fidelity Quantization}
\STATE Compute Composed Query Feature $\mathbf{F}_c$ and Target Feature $\mathbf{F}_t$.
\STATE Compute Similarity $\mathbf{S}$ and Quantify Fidelity $\mathcal{F}(\mathbf{F}_c, \mathbf{F}_t)$.
\STATE Separate batch into $\mathcal{T}_{{clean}}$ ($\mathcal{F} \ge \omega$) and $\mathcal{T}_{{noisy}}$ ($\mathcal{F} < \omega$).
\IF{$epoch \le N$}
\STATE // \textbf{2. Phase I: Negative Boundary Learning (NBL)}
\STATE Compute Diagonal Negative Composition feature $\mathbf{F}_{{neg}}$.
\STATE Compute Robust Contrastive Loss $\mathcal{L}_{{robust}}$.
\STATE Compute Query-oriented loss($\mathcal{L}_{{intra}}$) and Target-oriented ($\mathcal{L}_{{inter}}$).
\STATE Compute NBL Joint Objective: $\mathcal{L}_{{NBL}} = \mathcal{L}_{{robust}} + \zeta \mathcal{L}_{{intra}} + \nu \mathcal{L}_{{inter}}$.
\STATE Update model parameters $\Theta$ using $\nabla_{\Theta} \mathcal{L}_{{NBL}}$.
\ELSE
\STATE // \textbf{3. Phase II: Boundary-based Targeted Unlearning (BTU)}
\STATE // \textit{Optimal Transport for Unlearning}
\STATE Compute Cost Matrix $\mathbf{C}$ and Mask Matrix $\mathbf{M}$ using features $\mathbf{F}_c, \mathbf{F}_t, \mathbf{F}_{{neg}}$.
\STATE Solve Masked Entropy-Regularized OT problem for optimal transport plan $\mathbf{P}^*$.
\STATE Construct Smooth Soft Label $\mathbf{Y}$ (using $\mathbf{P}^*$ and Hard Label $\mathbf{L}$).
\STATE Compute Targeted Unlearning Loss $\mathcal{L}_{ul}$ (using ${KL}({Logit Matrix} || \mathbf{Y})$).
\STATE // \textit{Joint Optimization}
\STATE Compute Robust Contrastive Loss $\mathcal{L}_{{robust}}$ and $\mathcal{L}_{{intra}}$.
\STATE Compute BTU Joint Objective: $\Psi^* = \mathcal{L}_{{robust}} + \kappa \mathcal{L}_{ul} + \zeta \mathcal{L}_{{intra}}$.
\STATE Update model parameters $\Psi$ using $\nabla_{\Theta} \mathcal{L}_{{BTU}}$.
\ENDIF
\ENDFOR
\ENDFOR
\STATE \textbf{Return} Trained model parameters $\Psi^{*}$
\end{algorithmic}
\end{algorithm}

In the case of CIRR (bottom row), the reference image shows two monkeys, and the text instruction is ``more monkeys''. The retrieval results of $\mathbf{F}_c$ correctly point to images of monkey groups containing multiple monkeys, which reflects an understanding of the semantics regarding quantity increase. However, the retrieval results of $\mathbf{F}_{neg}$ show single monkeys or very few monkeys, which directly constitutes a semantic reversal of the instruction ``more'' in terms of quantity relationships. This phenomenon further confirms that $\mathbf{F}_{neg}$ is not a random vector in the feature space but a structured ``reverse signpost'' with explicit semantic direction. By explicitly constructing and utilizing it as a negative boundary, \modelname employs the Boundary-based Targeted Unlearning (BTU) module to define a clear boundary within the continuous metric space. Consequently, this achieves precise repulsion and unlearning of noise patterns without damaging neighboring clean samples.

\subsection{NTC Identification Analysis}
\label{sup:qualitative_NTC}

\begin{figure*}
    \centering
    \includegraphics[width=1\linewidth]{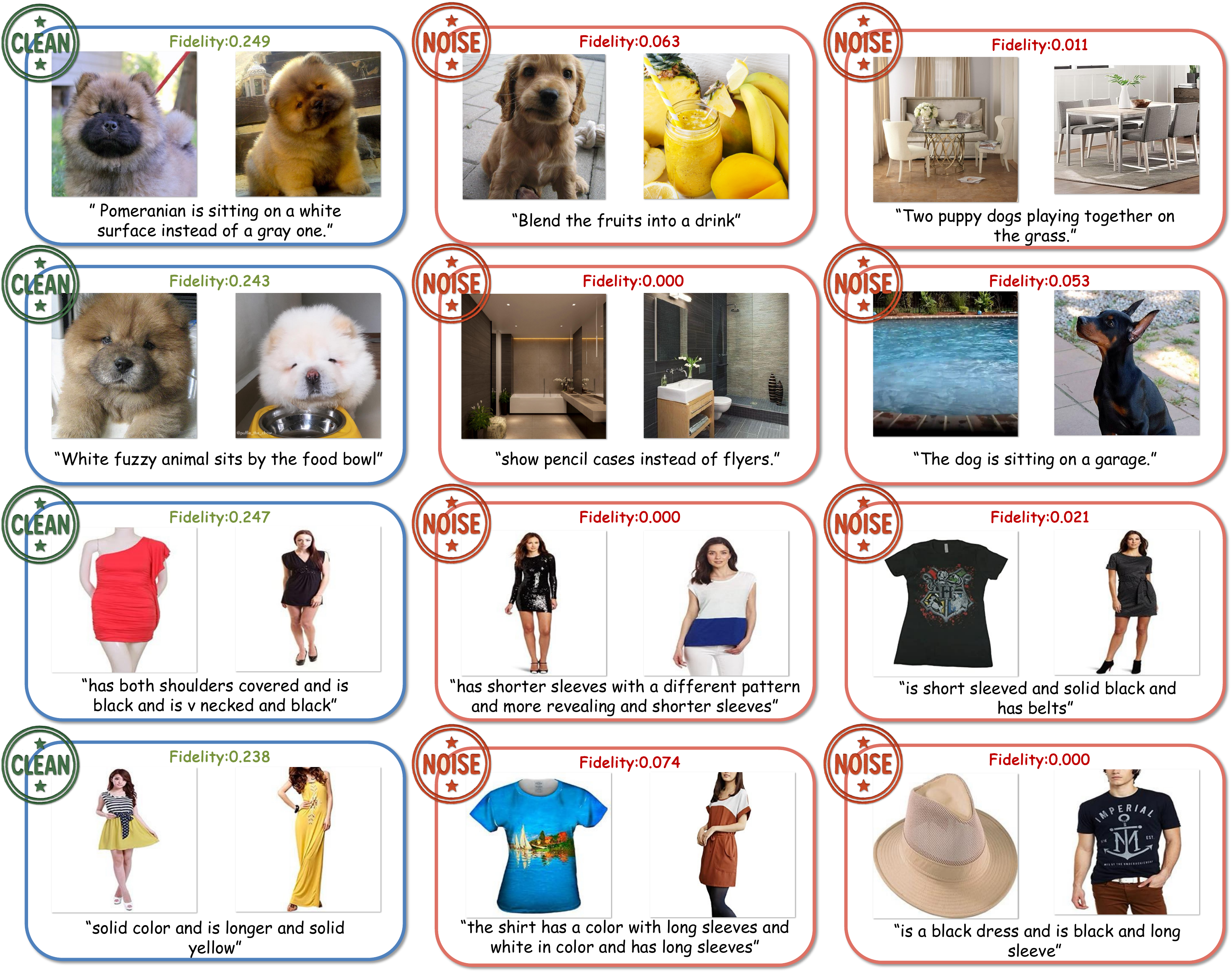}
\caption{\textbf{Visualization of NTC Identification Analysis.} We display the discrimination between \textcolor[HTML]{76943B}{\textbf{Clean}} (Left) and \textcolor[HTML]{B84F31}{\textbf{Noisy}} (Middle/Right) triplets by \modelname in the NTC scenario, along with the \textbf{Fidelity} scores computed by the \textbf{Geometric Fidelity Quantization (GFQ)} module. \modelname successfully distinguishes clean samples (assigned high fidelity scores, e.g., 0.249) from noisy ones. Notably, it effectively overcomes \textbf{Modality Suppression} in ``Hard Noise'' cases (e.g., Top-Right: visually similar rooms but mismatched text ``Two puppy dogs''), assigning them extremely low fidelity scores (e.g., 0.011). This precise geometric differentiation provides high-quality signals for the subsequent \textbf{Boundary-based Targeted Unlearning (BTU)}.}
    \label{fig:ntc_case}
\vspace{-12pt}
\end{figure*}

%%%%%%%%%%%%%%%%%%%%%%%%%%%%%%%%%%%%%%%%%%%%%
As shown in Figure~\ref{fig:ntc_case}, we present the discrimination of \modelname on Clean and Noisy triplets under NTC scenarios, along with the Fidelity scores calculated by the Geometric Fidelity Quantization (GFQ) module. The results demonstrate that \modelname is able to accurately distinguish between clean and noisy matches, outputting reasonable fidelity estimates, thereby exhibiting strong discriminative capability in overcoming Modality Suppression.

Specifically, the green area on the left highlights triplets identified as Clean by \modelname, which are generally assigned high fidelity scores. For example, in the top-left case, both the reference and target images depict fluffy dogs, with the modification text ``Pomeranian is sitting on a white surface instead of a gray one''. \modelname assigns a high fidelity estimation of 0.249, accurately reflecting that it captures the fine-grained semantic change of the background surface without being confused by the visual similarity of the dogs. Similarly, in the third row on the left, for the query ``has both shoulders covered and is black and is v necked'', \modelname correctly associates the reference red dress with the target black dress, assigning a high score of 0.247, indicating precise alignment with complex attribute modifications.

In contrast, the red areas in the middle and right columns display Noisy Correspondence, which are given noticeably lower fidelity scores. For instance, the example at the top of the middle column shows a reference image of a dog and a target image of a smoothie, with the text ``Blend the fruits into a drink''. This represents a clear cross-category semantic mismatch, and \modelname correctly classifies it as noise, with a fidelity of only 0.063. More importantly, \modelname effectively identifies Hard Noise where visual similarity might mask semantic inconsistency. In the top-right example, both the reference and target images depict indoor room scenes (high visual similarity), but the text describes ``Two puppy dogs playing together''. Despite the strong visual correlation between the images, \modelname successfully overcomes the Modality Suppression caused by visual dominance, assigning an extremely low fidelity score of 0.011 due to the complete textual misalignment.

Overall, this visualization strongly supports \modelname's capability for accurate Geometric Fidelity Quantization of triplets in NTC settings. \modelname not only identifies semantically consistent samples but also reliably assigns low scores to various types of noise (including subtle Hard Noise) by explicitly locating the geometric noise boundary. Such fine-grained geometric discrimination effectively provides high-quality signals for the subsequent Boundary-based Targeted Unlearning (BTU), thereby significantly enhancing the model's robustness.
%%%%%%%%%%%%%%%%%%%%%%%%%%%%%%%%%%%%%%%%%%%%%

\subsection{More Case Study}
\label{sup:qualitative_case}

\begin{figure*}[ht]
    \centering
    \resizebox{\linewidth}{!}{%
    \includegraphics[width=\linewidth]{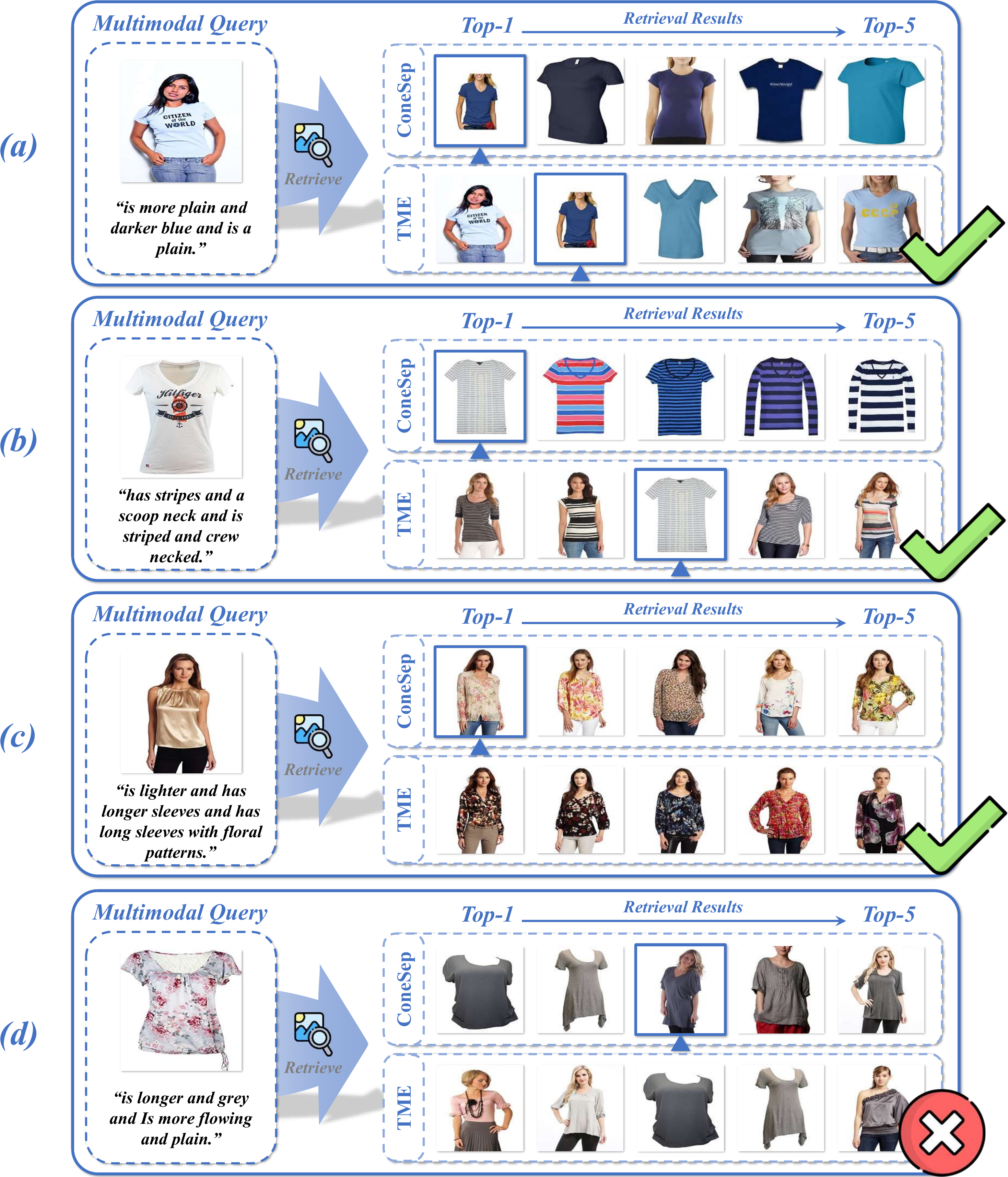}
    }
    \caption{Additional retrieval comparisons on \textbf{FashionIQ}. \modelname accurately follows fine-grained attribute changes (e.g., patterns, sleeve lengths), while TME often suffers from visual inertia from the reference image.}
    \label{fig:sup_more_fiq}
\end{figure*}

\begin{figure*}[ht]
    \centering
    \resizebox{\linewidth}{!}{%
    \includegraphics[width=\linewidth]{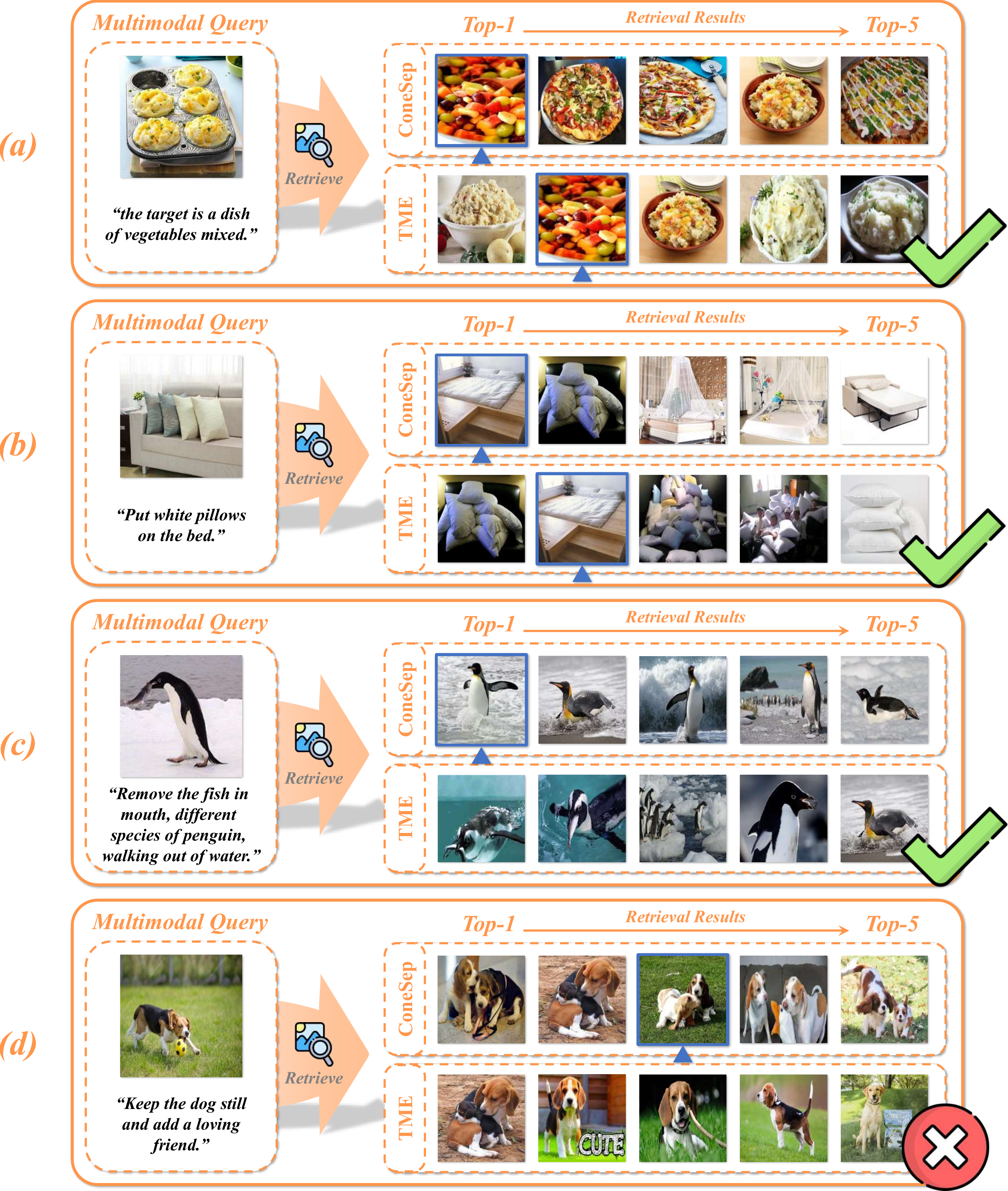}
    }
    \caption{Additional retrieval comparisons on \textbf{CIRR}. \modelname demonstrates superior capability in handling large semantic shifts (e.g., Muffin $\rightarrow$ Vegetable) and complex spatial/action modifications, whereas TME struggles to break away from the reference image's visual dominance.}
    \label{fig:sup_more_cirr}
\end{figure*}

To further comprehensively evaluate the effectiveness of \modelname in handling Noisy Triplet Correspondence (NTC) and complex semantic modification tasks, this section provides extended qualitative case studies on the FashionIQ and CIRR benchmark datasets. We present a comparison of the Top-5 retrieval results between \modelname and the current SOTA robust baseline, TME, in~\Cref{fig:sup_more_fiq} and~\Cref{fig:sup_more_cirr}. The images outlined in \textcolor{blue}{blue} represent the ground-truth (GT) targets, whose recall rankings intuitively reflect the retrieval accuracy and robustness of the models.

\statement{Analysis of Success Cases.}
\Cref{fig:sup_more_fiq} illustrates the retrieval results on the FashionIQ dataset, which emphasizes fine-grained attribute modifications. The comparative results clearly reveal the significant advantage of \modelname in overcoming ``Modality Suppression''. In cases (a) and (b), the user instructions require changing the color (to \textit{dark blue}) or the pattern (to \textit{striped}) of the reference image. The TME model exhibits strong ``visual inertia'', with its retrieval results often retaining the textual logo or the original style of the reference image, leading to semantic mismatch. In contrast, \modelname precisely decouples visual features from textual instructions, retrieving targets that match descriptions like ``\textit{solid dark blue}'' or ``\textit{striped scoop neck}''. Furthermore, when facing the complex multi-attribute instruction in case (c) involving ``\textit{lighter color}'', ``\textit{long sleeves}'', and ``\textit{floral patterns}'', \modelname demonstrates exceptional compositional reasoning capability. It accurately hits the target that satisfies all constraints simultaneously, whereas TME remains dominated by the dark visual features of the reference image, failing to effectively execute the color transformation instruction.

Similarly,~\Cref{fig:sup_more_cirr} presents results on the CIRR dataset, which involves more drastic semantic spans and spatial relationship reasoning. In case (a), the query requests changing ``\textit{muffins}'' to a ``\textit{vegetable platter}''. Due to the immense difference in visual features, traditional methods are prone to failure. TME retrieves mashed potatoes visually similar to muffins at the Top-1 rank, remaining trapped by the visual appearance of the reference image; conversely, \modelname successfully bridges the semantic gap, retrieving the correct vegetable category. In cases (b) and (c), whether it is the object category transition from ``\textit{sofa}'' to ``\textit{bed}'', or the dynamic action modification from ``\textit{holding fish}'' to ``\textit{walking out of water}'', \modelname accurately captures the core semantic changes. This further confirms that our proposed \textit{Geometric Fidelity Quantization (GFQ)} strategy can effectively identify and suppress hard noise interference that is visually similar but semantically mismatched, thereby guiding the model to focus on critical semantic information within the modification text.

\statement{In-depth Analysis of Failure Cases.}
Although \modelname exhibits strong robustness, retrieval misses can still occur in certain extreme scenarios. We conducted an in-depth analysis of typical ``failure cases'' in~\Cref{fig:sup_more_fiq}(d) and~\Cref{fig:sup_more_cirr}(d) and found that these cases actually reflect the ``False Negative'' challenge present in the current CIR evaluation system, which conversely corroborates the strong semantic understanding capability of \modelname.

Specifically, in case (d) of FashionIQ, the instruction requires changing a ``\textit{floral top}'' to one that is ``\textit{longer, grey, flowing, and plain}''. Although \modelname failed to recall the GT at Top-1 (the GT is at Top-3), observing its Top-1 and Top-2 results reveals that they are perfect ``\textit{grey plain long tops}'', which fully comply with the text description visually and semantically. In contrast, TME's results are mixed with garments containing pink patterns, clearly suffering from residual interference from the reference image. Similarly, in case (d) of CIRR, the core instruction is ``\textit{add a loving friend}'' (i.e., adding a dog). The Top-5 results of \modelname all display scenes of ``\textit{two dogs}'' interacting, demonstrating extremely high semantic consistency. However, TME's results waver between ``\textit{single dog}'' and ``\textit{two dogs}'', indicating its failure to stably capture the instruction regarding quantity change.

In summary, these so-called ``failure cases'' reveal that \modelname actually generates retrieval results that are more aligned with human intent than the Baseline. The semantic consistency demonstrated by the model in the Top-k candidate list is attributed to the negative boundary constructed by the \textit{Negative Boundary Learning (NBL)} module and the targeted unlearning mechanism of the \textit{Boundary-based Targeted Unlearning (BTU)} module. This enables the model to firmly ``push away'' erroneous visual features from the reference image (e.g., floral patterns, single dog) and precisely align with the user's true modification intent. This not only proves the robustness of \modelname in dealing with semantic uncertainty but also suggests directions for improvement in CIR dataset annotation and evaluation metrics for future work.

\clearpage
{
    \small
    \bibliographystyle{ieeenat_fullname}
    \bibliography{main}
}

% \end{CJK}
\end{document}